%% file: main.tex
\documentclass[letterpaper]{article} 
\usepackage[]{aaai23}  
\usepackage{times}  
\usepackage{helvet}  
\usepackage{courier}  
\usepackage[hyphens]{url}  
\usepackage{graphicx} 
\usepackage{natbib}  
\usepackage{caption} 
\usepackage{algorithm}
\usepackage{algorithmic}

\usepackage{newfloat}
\usepackage{listings}
\DeclareCaptionStyle{ruled}{labelfont=normalfont,labelsep=colon,strut=off} 
\floatstyle{ruled}
\newfloat{listing}{tb}{lst}{}
\floatname{listing}{Listing}
\def\arXiv#1{#1}

\usepackage{amsmath}
\usepackage{amssymb}
\usepackage{mathtools}
\usepackage{amsthm}
\usepackage{dsfont}
\usepackage{nicefrac} 
\usepackage{enumitem}
\usepackage{csquotes}

\usepackage[capitalize,noabbrev]{cleveref}

\usepackage{standalone}
\usepackage{tikz, pgf}
\usepackage{pgfplots}
\usepackage{pgfplotstable}
\usepackage{tikz-network}

\theoremstyle{plain}

\theoremstyle{definition}

\theoremstyle{remark}

\renewcommand{\vec}[1]{\ensuremath{\boldsymbol{#1}}}

\newcommand{\mat}[1]{{\ensuremath{{\mathbf{#1}}}}}

\def\DEF{\triangleq} 
\def\T{^\mathrm{T}} 
\newcommand{\dd}{\operatorname{d}\!}
\DeclareMathOperator{\E}{E}
\DeclareMathOperator{\diag}{diag}
\DeclareMathOperator{\Tr}{Tr}
\newcommand{\NewR}{\ensuremath{\mathds{R}}}

\def\va{\vec a}
\def\vmu{\vec \mu}
\def\vs{\vec s}
\def\vt{\vec t}
\def\vw{\vec w}
\def\vx{\vec x}
\def\vy{\vec y}
\def\vz{\vec z}

\def\C{\mat C}
\def\W{\mat W}
\def\X{\mat X}
\def\Y{\mat Y}

\def\J{\mat J}
\def\L{\mat L} 

\def\SN{\mathcal N}
\def\SD{\mathcal D}
\def\SW{\mathcal W}

\title{Kalman Bayesian Neural Networks for Closed-form Online Learning}
\author{
    Philipp Wagner\textsuperscript{\rm 1}\equalcontrib,
    Xinyang Wu\textsuperscript{\rm 1}\equalcontrib,
    Marco F. Huber\textsuperscript{\rm 1,2}
}
\affiliations{
    \textsuperscript{\rm 1}Department Cyber Cognitive Intelligence (CCI), Fraunhofer Institute for Manufacturing Engineering and Automation IPA, Stuttgart, Germany
		
		\textsuperscript{\rm 2}Institute of Industrial Manufacturing and Management IFF, University of Stuttgart, Germany\\

    philipp.wagner@ipa.fraunhofer.de, xinyang.wu@ipa.fraunhofer.de, marco.huber@ieee.org
}

\usepackage{bibentry}

\begin{document}

\maketitle

\begin{abstract}
Compared to point estimates calculated by standard neural networks, Bayesian neural networks (BNN) provide probability distributions over the output predictions and model parameters, i.e., the weights. Training the weight distribution of a BNN, however, is more involved due to the intractability of the underlying Bayesian inference problem and thus, requires efficient approximations. In this paper, we propose a novel approach for BNN learning via closed-form Bayesian inference. For this purpose, the calculation of the predictive distribution of the output and the update of the weight distribution are treated as Bayesian filtering and smoothing problems, where the weights are modeled as Gaussian random variables. This allows closed-form expressions for training the network's parameters in a sequential/online fashion without gradient descent. We demonstrate our method on several UCI datasets and compare it to the state of the art.
\end{abstract}

\section{Introduction} \label{ch:sec1}
Deep Learning has been continuously attracting researchers for its applicability in many fields such as medical diagnostics \cite{amisha2019overview}, autonomous control \cite{zeng2020tossingbot}, or intelligent mass-productions \cite{el2019simulation}. However, conventional deep Neural Networks (NNs) usually perform maximum likelihood estimation, which results solely in a point estimate without consideration of uncertainty in the data and the learned model. In domains with high safety standards or financial risks this approach is not sufficient and limits the number of possible applications. Bayesian methods offer ways to overcome this issue by quantifying uncertainties using Bayes' rule and probabilistic reasoning, which results in a distribution over network parameters and predictions instead of point estimates. A quantification of the uncertainty indicates whether the predictions are trustworthy and reliable \cite{begoli2019need}. Popular approaches like Markov Chain Monte Carlo (MCMC) are computationally demanding, whereas variational inference (VI) or ensemble methods rely on noisy gradient computations and need to be trained using batched training data and several iterations. Due to these characteristics, the mentioned methods are not directly applicable in online learning settings \cite{parisi2019}, but first approaches using a data memory exist \cite{Nguyen2018, kurle2019continual}. In addition, gradient-based methods may suffer from poor choices of the optimization hyper-parameters or bad local minima \cite{Bengio2012}. This behavior is mitigated by adaptive learning rates, stochastic gradient descent (SGD), and modern optimizers, but still persists.

In this paper we develop a novel online learning approach for Bayesian Neural Networks (BNN) \cite{mackay1992practical} that is named \emph{Kalman Bayesian Neural Network (KBNN)}. The key idea is to train the BNN via sequential Bayesian filtering without the need of gradient-based optimization. Bayesian filtering is commonly used to estimate probability density functions (PDF) from noisy observations in Markov chains \cite{saerkkae_2013, huber2015nonlinear}. Many NN architectures like the one studied in this paper also form a Markov chain \cite{Achille2018}. Our approach is motivated by the work of \citet{puskorius2001parameter}, in which the training procedure is also treated as a filtering problem, which however uses local linearization based on backpropagated gradient information to compute weight updates. While the special case of online training of a single perceptron is discussed by \citet{huber2020bayesian}, we aim at constructing a learning algorithm for a universal multilayer perceptron (MLP). 

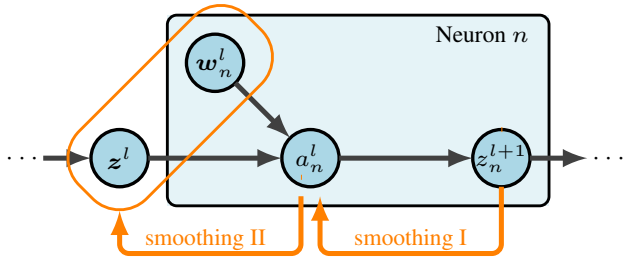
\begin{figure}[t]
\centering
\resizebox{0.48\textwidth}{!}{\input{figures/PGM/PGM_tikz2.tex}}
\caption{Probabilistic graphical model for an arbitrary layer $l=1\ldots L$. During the backward pass, first $a_n^l$ is updated via a Kalman smoothing step and afterwards $\vz^{l}$ and $\vw_n^l$ are updated jointly again via smoothing. For the output layer $l=L$ the ground truth $\vy$ is known from the data $\SD$ and thus, for updating the output $\vz^{l+1}$ a filtering step instead of a smoothing step is performed. Smoothing I refers to \eqref{eq:smootheda} while smoothing II refers to \eqref{eq:muwupdate} and \eqref{eq:cwupdate}.}
\label{fig:backwar}
\end{figure}

The KBNN consists of a forward pass for inferring the output distribution given an input sample and a backward pass to adjust the network parameters given a corresponding output sample. A part of the backward pass is visualized in Fig.~\ref{fig:backwar}. By assuming that the weights and outputs of each network layer are Gaussian distributed, it is sufficient to compute the mean and covariance in a moment matching fashion. In doing so, it is possible to provide closed-form expressions for the forward pass as well as the update equations of the weights in the backward pass for commonly used activation functions like sigmoid or ReLU. The main contributions of this paper are: 
(a) We introduce an approach that strictly follows Bayesian inference. Thus, learning the network parameters is not reformulated as optimization problem. The use of Bayesian filtering and smoothing techniques instead enables native online training of BNNs, where we show the relationship to the famous Kalman filtering and smoothing equations \cite{kalman1960filter, rauch1965smoother}.
(b) We extend the work of \citet{huber2020bayesian} being applicable only for a single neuron to the case of an MLP. In doing so, all calculations necessary are performed in closed form. For ReLU and linear activations they are exact.
(c) We compare the performance of our approach on various common benchmarking datasets to inference techniques such as MCMC, VI and expectation propagation (EP).


\section{Related Work}
\label{sec:relatedwork}
%
\paragraph{Laplace Approximation}%
The idea of Bayesian inference in the NN setting goes back to the work by \citet{mackay1992practical}, where a probability distribution over the model parameters is learned via Laplace transformation. Here, a Gaussian distribution is fitted to the vicinity of the maximum posterior estimate. Due to that point of time, Bayesian inference was limited to small networks and datasets. 
This approximation technique recently gained increased interest by \cite{ritter2018laplace, hennig2020bayesian}. \citet{Snoek2015, hennig2020bayesian} for instance use it to provide uncertainty estimates only for the last layer of an MLP.

\paragraph{MCMC}
One of the most explored ideas for probabilistic inference in general and for learning BNNs in particular is MCMC \cite{Metropolis1953equationOS}, which allows approximating probability integrals with the Monte Carlo method via sampling from a Markov process. Many improvements have been suggested for the initial Metropolis-Hastings algorithm such as Gibbs sampling \cite{geman1984gibbssampling}, hybrid Monte Carlo \cite{neal1995bnn}, or Hamiltonian Monte Carlo (HMC) \cite{DUANE1987hmc}. An important extension to HMC is the No-U-Turn Sampler (NUTS)~\cite{hoffman2014no}, which mostly performs more efficiently. One downside of most MCMC approaches is the high computational cost necessary for performing density estimation. Further, only samples and no closed-form representation of the distribution are available.

\paragraph{Variational Inference} 
The usage of VI for BNN training was introduced by \citet{graves2011practical}. VI is based on the idea of approximating the complicated weight posterior by means of a simple distribution like a Gaussian. This approximation is achieved by minimizing the empirical lower bound to the reverse Kullback-Leibler divergence using gradient descent. 
In \citet{Kingma2014AutoEncodingVB}, the gradient is estimated via sampling leading to high variance gradients and thus, merely a decent performance only in smaller architectures. In \citet{JMLR:v14:hoffman13a} a more scalable method called Stochastic Variational Inference (SVI) is proposed, which computes a scaled gradient on randomly sampled subsets of data to update the variational parameters, instead of computing gradients from the full dataset. A deterministic calculation is proposed in \citet{wu2019}, which uses a closed-form forward pass as in our work. 
\citet{gal16dropout} found that the dropout technique introduced by \citet{srivastava2014dropout} approximates the variational distribution while being relatively computationally cheap. 

\paragraph{Expectation Propagation}
Minimizing the forward Kullback-Leibler divergence instead of its reverse version leads to EP \cite{minka2001}. In contrast to VI, EP is not guaranteed to convergence in general. A practical EP version for BNNs named probabilistic backpropagation (PBP) was proposed in \citet{hernandez2015probabilistic} and extended in \citet{Ghosh2016}, which share similarities with our work. The forward pass of PBP also utilizes closed-form moment propagation. However, \citet{Ghosh2016} and \citet{hernandez2015probabilistic} employ the mean-field approximation, i.e., fully factorized Gaussian distributions, while we allow correlated weights per neuron. Significant difference are given for the backward pass. Here, the KBNN requires no explicit calculation of the marginal likelihood and its first and second order derivatives. 


\paragraph{Kalman Filtering}
Closest to our work is the usage of Kalman filtering for training BNNs. One of the first approaches was proposed by \citet{watanabe1990}, where the weights of the networks are assumed to be Gaussian. The mean and variance of the individual weights are updated by means of an extended Kalman filter, which however requires local linearization for updating the hidden neurons. This work was extended by \citet{puskorius2001parameter} to allow layer-wise correlated or even network-wide correlated neurons. 
To avoid linearization, \citet{huber2020bayesian} proposes the so-called Bayesian perceptron. Even though limited to a single neuron, this work shows that closed-form Bayesian inference for calculating the mean and covariance parameters of the weight posterior distribution is possible. In the following, we extend this single neuron approach to an MLP by utilizing Bayesian filtering and smoothing.

\paragraph{Online Learning}

In the context of online learning Bayesian methods are a popular choice, since uncertainties over the data and the model can be taken into account directly. \citet{Opper1998} use a moment matching approach for online learning which is similar to our work. \citet{kirkpatrick2017forgetting} and \citet{ritter2018online} deal with the issue of catastrophic forgetting in neural networks for continual learning tasks. There are a few works that include data memory to improve online learning capabilities \cite{minka2009vvm,Nguyen2018,kurle2019continual}.

\section{Problem Formulation}
\label{sec:problem}
Given a dataset $\SD = \{ (\vx_i, \vy_i) \}_{i=1}^N$ of $N$ i.i.d. pairs of training instances with inputs $\vx_i \in \NewR^d$ and outputs $\vy_i \in \NewR^e$, we want to train an MLP with $L$ layers in a supervised learning setup. In each layer $l = 1\ldots L$, a nonlinear transformation 
\begin{equation}
	\label{eq:activation}
   \vz^{l+1} = f(\va^{l})  \text{ with }\, \va^{l} = \W^{l} \cdot \vz^{l} + \vw_0^{l} 
\end{equation}
is performed with weight matrix $\W^{l} \DEF [\vw_1\,\ldots\,\vw_{M_l}]\T\in\NewR^{M_{l}\times M_{l-1}}$ with $\vw_i \in \NewR^{M_{l-1}}$ being the $i$-th neuron's weight vector, bias $\vw_0^{l}\in\NewR^{M_l}$, and nonlinear activation function $f(\cdot)$, where $M_l$ is the number of neurons of the $l$-th layer. The output $\vz^{l+1}$ of layer $l$ becomes the input of the subsequent layer $l+1$. For the first layer $\vz^{1} = \vx$ and for the last layer $\vz^{L+1} = \vy$. To simplify the notation, we avoid the layer superscript $l$ for $\va$ and $\W$ whenever possible.

By redefining the input $\vz^l \DEF [1\, z_1^l\, \ldots\, z_{M_{l-1}}^l]\T$ we can conveniently incorporate the bias $\vw_0$ into the weights according to $\vw_i \DEF [w_0^i\, w_1^i\, \ldots\, w_{M_{l-1}}^i]\T$ where $w_0^i$ is the $i$-th element of $\vw_0$ and thus, $\W \in \NewR^{M_{l}\times (M_{l-1}+1)}$. Further, $\SW\DEF\{\W^l\}_{l=1}^L$ comprises all weight matrices. Whenever appropriate, we use $\vw \DEF [\vw_1\T\,\ldots\,\vw_{M_l}\T]\T = \operatorname{vec}(\W)$ to simplify the calculations and notation.

The MLP is treated as a BNN. Accordingly, the weights in $\SW$ are random variables with (prior) probability distribution $p(\SW)$. The task now is two-fold \cite{mackay1992practical, neal1995bnn}: (i) Calculating the posterior distribution of the weights
\begin{equation}
    \label{eq:bayes}
    p(\SW|\SD) = \frac{p(\Y|\X, \SW)\cdot p(\SW)}{p(\Y|\X)}~,
\end{equation}
with normalization term $p(\Y|\X)$ and $\X \DEF [\vx_1\, \ldots\, \vx_N]$, $\Y \DEF [\vy_1\, \ldots\, \vy_N]$ being the input and output data from~$\SD$, respectively. (ii) Calculating the predictive distribution
\begin{equation}
    \label{eq:prediction}
    p(\vy|\vx, \SD) = \int p(\vy|\vx,\SW)\cdot p(\SW|\SD)\dd\SW
\end{equation}
of the BNN given a new input $\vx$. Unfortunately, both equations cannot be solved exactly in closed form in general. To provide an approximate but closed-form solution we employ techniques from \emph{Bayesian filtering and smoothing}, which is usually applied to Markov processes in order to estimate a state variable over time from noisy observations. Equation \eqref{eq:activation} forms a continuous-valued Markov chain with random variables $\va$, $\W$, and $\vz$ as depicted in the graphical model Fig.~\ref{fig:backwar}. Hence, \eqref{eq:prediction} can be solved layer-wise by means of consecutive prediction steps of a Bayesian filter and will be addressed in the \emph{forward pass} of the proposed KBNN. Solving \eqref{eq:bayes} to train the weights requires filtering and smoothing steps of a Bayesian filter and is covered by the \emph{backward pass} of the KBNN. To obtain closed-form expressions in both passes, we make two key assumptions.

\paragraph{Assumption 1}
\label{as:independent} For BNNs it is very common to apply the strict version of the mean-field approximation, where all weights are assumed to be independent. In this paper, it is merely assumed that all neurons are pair-wise independent, so that the weights of individual neurons are dependent. 
This assumption significantly simplifies the calculations. The implications of dependent neurons are discussed in \citet{puskorius2001parameter} and Sec.~\ref{sec:discussion}.

\paragraph{Assumption 2}
\label{as:gaussian} For a single neuron, the corresponding quantities in \eqref{eq:activation} are assumed to be jointly Gaussian distributed. 
    Due to this assumption, particularly the posterior in \eqref{eq:bayes} and the predictive distribution in \eqref{eq:prediction} are approximated by means of Gaussian distributions and thus, our approach becomes an \emph{assumed density filter} \cite{Maybeck1979, Opper1998}. In doing so, it is sufficient to calculate the first two moments (mean and covariance) of the posterior and predictive distribution. For ReLU activations this approach even transforms into \emph{moment matching}.

Based on these assumptions, the posterior weight distribution of a layer is given in factorized form $p(\W|\SD) = \prod_{i=1}^{M_l} \SN(\vw_i|\vmu_{w}^i, \C_w^i)$, where $\SN(\vx|\vmu_x, \C_x)$ is a Gaussian PDF with mean vector $\vmu_x$ and covariance matrix~$\C_x$. 
The limitations arising from these assumptions are discussed in greater detail in the Sec.~\ref{sec:discussion}.

\section{The Kalman Bayesian Neural Network}
\label{sec:kbnn}
For deriving the forward and backward pass of the KBNN we process each training data instance $(\vx_i, \vy_i)$ individually and sequentially. This is possible as the data is assumed to be i.i.d. and thus, the likelihood in \eqref{eq:bayes} can be factorized according to $p(\Y|\X, \SW) = \prod_i p(\vy_i|\vx_i, \SW)$. Hence, we obtain the posterior $p(\SW|\SD_i) \propto  p(\vy_i|\vx_i, \SW)\cdot p(\SW|\SD_{i-1})$, with $\SD_i = \{(\vx_j, \vy_j)\}_{j=1}^i \subset \SD$ and $p(\SW|\SD)\equiv p(\SW|\SD_N)$, by means of recursively processing the data instances, where the recursion commences from the prior $p(\SW)$. During each recursion step it is not necessary to update all weights of the BNN simultaneously. Instead, we can make use of the Markov chain characteristic of the network (cf. Fig.~\ref{fig:backwar}). In doing so, the weight matrix of each layer can be updated one after the other. This updating is essentially performed during the backward pass, but requires intermediate predictive quantities $\va$ and $\vz$ that are calculated during the forward pass. Thus, the forward pass is not only necessary to calculate the predictive distribution of the BNN for new inputs, but is also a crucial component of the backward pass. Hence, we start with deriving the forward pass, where we omit the quantity $\SD$ in the following to simplify the notation.

\subsection{Forward Pass}
\label{sec:kbnn_forward}
During the forward pass the predictive distribution $p(\vy|\vx, \SD)$ for a given input $\vx$ has to be computed. For this purpose, information is propagated forward through the BNN in order to calculate the predictive distributions of all random variables $\va$ and $\vz$ along the path from the input to the output. Since these two quantities occur in each layer with the same dependencies, we restrict the derivation to a single layer without loss of generality. For the $l$-th layer the predictive distribution of the output $\vz^{l+1}$ is given by
\begin{equation}
    \label{eq:pdfforward}
    p(\vz^{l+1}|\vx) = \int p(\vz^{l+1}|\va) \cdot  p(\va|\vx) \dd \va
\end{equation}
with
\begin{equation}
    \label{eq:pdfforwarda}
    p(\va|\vx) = \int p(\va|\vz^{l}, \W) \cdot p(\vz^{l}|\vx) \cdot p(\W) \dd \vz^{l} \dd\W. 
\end{equation}
All quantities in \eqref{eq:pdfforwarda} are related according to \eqref{eq:activation}. Further, $p(\vz^{l}|\vx)$ and $p(\W)$ are assumed to be Gaussian. The predictive distribution $p(\va|\vx)$ however, is not Gaussian due to the multiplicative nonlinearity in \eqref{eq:activation}, but it is approximated by the Gaussian $\SN(\va|\vmu_a, \C_a)$ with mean vector and covariance matrix matching the moments of the true distribution. The elements of the mean vector $\vmu_a$ are given by
\begin{equation} 
    \label{eq:mua}
    \mu_{a}^n = \E[\vw_n\T\cdot \vz^l] = \E[\vw_n\T]\cdot \E[\vz^l] = (\vmu_{w}^n)\T \cdot \vmu_z^l~,
\end{equation}
while the covariance matrix is diagonal due to Assumption~1 with elements
\begin{align}
    (\sigma_{a}^n)^2 &= \E\hspace{-.5mm}\big[a_n^2\big] - (\mu_{a}^n)^2 
		= \E\hspace{-.5mm}\big[(\vw_n\T\cdot \vz^l)^2\big] - (\mu_{a}^n)^2 \hspace{10mm} \nonumber \\
    \label{eq:sigmaa}
    &=(\vmu_{w}^n)\T \C_z^l \vmu_{w}^n\hspace{-.5mm} + \hspace{-.5mm}(\vmu_z^l)\T \C_{w}^n \vmu_z^l\hspace{-.4mm} + \hspace{-.4mm}\Tr(\C_{w}^n \C_z^l)\,,
\end{align}
where $n=1\ldots M_l$ is the neuron index, $\Tr(\cdot)$ is the matrix trace, and $\vmu_z^l, \C_z^l$ are the mean and covariance of $\vz^l$. 
For the first layer $\vz^l = \vx$ and thus, no random variable. This allows solving \eqref{eq:pdfforwarda} exactly as $\va^l$ in \eqref{eq:activation} becomes a linear function, where $p(\va|\vx)$ is actually Gaussian. With $p(\vz^{l}|\vx) = \delta (\vz^{l} - \vx)$ in \eqref{eq:pdfforwarda} the means \eqref{eq:mua} and variances \eqref{eq:sigmaa} become $\mu_a^n = \vx\T\cdot \vmu_w^n$ and $(\sigma_a^n)^2 = \vx\T\C_w^n\vx$, respectively, which corresponds to a \emph{Kalman prediction} step.

The predictive distribution $\vz^{l+1}$ in \eqref{eq:pdfforward} is also approximated with a Gaussian $\SN(\vz^{l+1}|\vmu_z^{l+1}, \C_z^{l+1})$, where the elements of the mean vector and (diagonal) covariance matrix are given by
\begin{align}
    \label{eq:muz}
    \mu_z^{l+1,n} &= \E[f(a_n)]~, \\
    \label{eq:varz}
    (\sigma_z^{l+1,n})^2 &= \E\left[f(a_n)^2\right] - (\mu_z^{l+1,n})^2~,
\end{align}
respectively, and thus depend on the nonlinear activation function. For ReLU activations, the expected values in \eqref{eq:muz} and \eqref{eq:varz} can be calculated exactly in closed form and thus, we obtain a moment matching approximation. For sigmoidal activations like sigmoid or hyperbolic tangent, the expected values can be tightly approximated in closed form, except for the special case of a probit activation, where we again obtain a moment matching. 
Detailed derivations for both activations can be found in \citet{huber2020bayesian}\arXiv{ and the supplementary material}.

The distribution $p(\vz^{l+1}|\vx)$ is then used for solving the corresponding integrals \eqref{eq:pdfforward} and \eqref{eq:pdfforwarda} of the subsequent layer $l+1$. For $l=L$, we have $\vz^{l+1} = \vy$ and thus $p(\vz^{l+1}|\vx)$ coincides with the desired predictive distribution $p(\vy|\vx, \SD)$. 
The complete forward pass is listed in Algorithm~\ref{alg:forward}. 
It is worth mentioning that the calculated moments of each layer must be stored, as they are needed for the weight update procedure during the backward pass.

\begin{algorithm}[tb]
   \caption{Forward Pass of the KBNN for a new input $\vx$.}
   \label{alg:forward}
\begin{algorithmic}[1]
    \STATE  $\left( \vmu^{1}_z, \C^{1}_z \right) \leftarrow \left( \vx, \mathbf 0 \right)$
    \FOR{$l=1$ to $L$}
    \STATE Calc. mean $\vmu_a^l$ and covariance $\C_a^l$ via \eqref{eq:mua} and \eqref{eq:sigmaa}
    \STATE Calc. mean $\vmu_z^{l+1}$ and covariance $\C_z^{l+1}$ via \eqref{eq:muz} and \eqref{eq:varz}
    \ENDFOR
    \STATE Return $ \left( \vmu_a^l, \C_a^l, \vmu_z^{l+1}, \C_z^{l+1}\right) $ for $l=1\ldots L$
\end{algorithmic}
\end{algorithm}

\subsection{Backward Pass}
\label{sec:kbnn_backward}
The training of conventional MLPs relies on a problem specific loss function being optimized with SGD, where the entire dataset $\SD$ is processed repeatedly. The backward pass of the KBNN updates the weights by means of sequentially processing the data once without gradient-based optimization thanks to strictly following Bayes' rule in \eqref{eq:bayes}. Like with the forward pass, the Markov property of the network allows updating the weights layer-wise. Given any training instance $(\vx, \vy)$,  updating the $l$-th layer requires considering joint Gaussian PDFs of the form
\begin{align}
    p(\vs,\vt|\vx, \vy) 
    &= \SN\hspace{-.5mm}\left(\vs,\vt\left|
    \begin{bmatrix}
        \vmu_s^+ \\ \vmu_t^+
    \end{bmatrix}, \right.\begin{bmatrix}
    \C_s^+ & \C_{st} \\ \C_{st}\T & \C_t^+
    \end{bmatrix}\right) \nonumber \\
    \label{eq:jointgaussian}
    &= p(\vs|\vt, \vx) \cdot p(\vt|\vx, \vy)
\end{align}
twice: (I) $\vs=a_n$, $\vt=\vz_n^{l+1}$ and (II) $\vs=\left[ \vw\T\, (\vz^{l})\T \right]\T$, $t=\va$ as the graphical model in Fig.~\ref{fig:backwar} indicates. Here, $\vw = \mathrm{vec}(\W)$ is the vectorized weight matrix as defined in Sec.~\ref{sec:problem}. The Gaussian $p(\vt|\vx, \vy) = \SN(\vt|\vmu_t^+, \C_t^+)$ is already known from the previous step, while the conditional Gaussian $p(\vs|\vt, \vx)=\SN(\vs|\vmu_s^- + \J\cdot(\vt-\vmu_t^-), \C_s^- - \J \cdot\C_{st}\T)$ with \emph{Kalman gain} $\J = \C_{st}\cdot(\C_t^-)^{-1}$ \cite{huber2015nonlinear, saerkkae_2013}. 
%
The superscript~$-$ indicates quantities $p(\cdot|\vx)$ of the forward pass, while $+$ is the updated version $p(\cdot|\vx, \vy)$ resulting from the backward pass. Calculating the product of the two Gaussian PDFs in \eqref{eq:jointgaussian} and marginalizing $\vt$ yields
\begin{align}
    \begin{split}
    \label{eq:smootheds}
    \vmu_s^+ &= \vmu_s^- + \J\cdot(\vmu_t^+ - \vmu_t^-)~, \\
    \C_s^+ &= \C_s^- + \J\cdot(\C_t^+ - \C_t^-)\cdot\J\T
    \end{split}
\end{align}
being the mean and covariance of $p(\vs|\vx,\vy)$, respectively. These equations correspond to the \emph{Kalman} or \emph{Rauch-Tung-Striebel smoother} \cite{rauch1965smoother}.

For the smoothing step (I), \eqref{eq:smootheds} becomes
\begin{align}
    \begin{split}
    \label{eq:smootheda}
    \mu_a^{n,+} &= \mu_a^{n,-} + \vec k_n\T\hspace{-.5mm}\cdot\hspace{-.3mm}(\vmu_{z}^{l+1,+} - \vmu_z^{l+1,-})~, \\ 
    (\sigma_a^{n,+})^2 &=  (\sigma_a^{n,-})^2 + \vec k_n\T\hspace{-.5mm}\cdot\hspace{-.3mm} (\C_{z}^{l+1,+}\hspace{-.5mm} - \C_{z}^{l+1,-})\hspace{-.5mm}\cdot\hspace{-.3mm}\vec k_n\,,
    \end{split}
\end{align}
for neuron $n=1\ldots M_l$ with $\vec k_n = (\C_z^{l+1,-})^{-1}\cdot (\vec \sigma_{az}^n)^2$. All quantities in \eqref{eq:smootheda} can be calculated in closed form but the cross-covariance $(\vec \sigma_{az}^n)^2$, which depends on the activation function. As with the forward pass, ReLU allows an analytical solution, while for sigmoidal activations a closed-form approximation exists. For details be referred to \citet{huber2020bayesian}\arXiv{ and the supplementary material}.

The result $\vs$ of step (I) becomes the quantity $\vt$ of step (II), for which the mean and covariance in \eqref{eq:smootheds} are given by
\begin{align}
    \label{eq:muwupdate}
    \begin{bmatrix} \vmu_{w}^{+} \\ \vmu_{z}^{l,+} \end{bmatrix} &=  \begin{bmatrix} \vmu_{w} \\ \vmu_{z}^{l,-} \end{bmatrix} + \L\cdot( \vmu_a^{+} - \vmu_a^{-} )~, \\
    \label{eq:cwupdate}
    \begin{bmatrix} \C_{w}^{+} & \C_{wz} \\ \C_{wz}\T & \C_{z}^{l,+} \end{bmatrix} &= 
    \C + \L\cdot(\C_a^{+} - \C_a^{-})\cdot \L\T, 
\end{align}
with $\L= \C_{wza}\cdot(\C_a^{-})^{-1}$, $\C = \diag(\C_{w}, \C_{z}^{l,-})$, and $\C_a^+ = \diag((\sigma_a^{1,+})^2,\ldots,(\sigma_a^{M_l,+})^2)$. 
The structure of the covariance can be explained as follows. At the forward pass, $\W$ and $\vz^l$ are independent as $\va$ is not observed and these three quantities are connected via a v-structure $\W \rightarrow \va \leftarrow \vz^l$ (cf. Fig.~\ref{fig:backwar}). Thus, $\C$ has a block-diagonal structure. At the backward pass, a descendent of $\vz^l$, namely $\vy$ is observed and thus, $\W$ and $\vz^l$ are dependent. 
The mean $\vmu_w^{+}$ and covariance $\C_w^{+}$ are the updated weight parameters of $p(\vw|\vx, \vy, \SD_{i-1})$, while $\vmu_z^{l,+}$ and $\C_z^{l,+}$ are used for the quantity $\vt$ of step (I) of layer $l-1$. 
This update rule differs from \citet{huber2020bayesian} since $\vz^{l}$ is not deterministic for any layer but the input layer. All quantities are known except of $\C_{wza}$, which is given by 
\begin{align}
    \C_{wza} 
    &= \E \left[ \left(\begin{bmatrix} \vw \\ \vz \end{bmatrix} - \begin{bmatrix} \vmu_w \\ \vmu_z^- \end{bmatrix}\right) \cdot (\va - \vmu_a^- )\T \right]  \nonumber \\
    &= \begin{bmatrix} \diag\hspace{-.5mm} \left( \C_{w}^{1} \cdot \vmu_z^{l,-}, \ldots , \C_{w}^{M_l} \cdot \vmu_z^{l,-} \right) \\
        \C_z^{l,-} \cdot \vmu_{w}^{1}\quad \cdots \quad \C_z^{l,-} \cdot \vmu_{w}^{M_l}
    \end{bmatrix}~. \label{eq:cwza}
\end{align}
The black-diagonal structure of the upper part of $\C_{wza} $ is due to Assumption~1. \arXiv{The detailed derivation can be found in the supplementary material.}

\begin{algorithm}[tb]
   \caption{Backward pass for training on dataset $\SD$}
   \label{alg:backward}
\begin{algorithmic}[1]
    \FOR{each training instance $( \vx_i, \vy_i ) \in \SD$}
    \STATE $\left( \vmu_a^{l,-}, \C_a^{l,-}, \vmu_z^{l+1,-}, \C_z^{l+1,-}\right) \leftarrow \mathrm{ForwardPass}(\vx_i)$
    \STATE $\left( \vmu_{z}^{L+1,+}, \C_{z}^{L+1,+} \right) \leftarrow \left( \vy_i, \mat 0 \right)$
    \FOR{$l=L$ to 1}
    \STATE Update $\vmu_{a}^+$, $\C_{a}^+ $ via \eqref{eq:smootheda}
    \STATE Update $\vmu_{w}^+$, $\C_{w}^+$, $\vmu_{z}^{l,+}$, $\C_{z}^{l,+}$ via \eqref{eq:muwupdate} and \eqref{eq:cwupdate}
    \STATE Store $\left( \vmu_w, \C_w \right) \leftarrow \left( \vmu_w^+, \C_w^+ \right)$
    \ENDFOR
    \ENDFOR
\end{algorithmic}
\end{algorithm}

The sequence of smoothing operations is not surprising as updating is not performed with the data directly but with previously estimated quantities. The only exception is layer $L$, where the recursion starts. Here, in step (I) $\vt\hspace{-.2mm}=\hspace{-.3mm}\vz^{L+1}\hspace{-.5mm}=\hspace{-.5mm}\vy$ is deterministic, thus $\vmu_z^{L+1,+}\hspace{-.5mm}=\hspace{-.5mm}\vy$~and $\C_{z}^{L+1,+}\hspace{-.5mm}=\hspace{-.5mm}\mat 0$.~By substituting these quantities in \eqref{eq:smootheda} the Kalman smoother becomes a (nonlinear) \emph{Kalman filter} \cite{kalman1960filter}. The backward pass is summarized in Algorithm~\ref{alg:backward}.

\section{Experiments}
\label{sec:experiments}
In this section, we validate the proposed KBNN in both classification and regression tasks on benchmark datasets. Four experiments are conducted: (i) Evaluating the KBNN on a synthetic regression task, (ii) binary classification on the well-known Moon dataset, (iii) online learning on the Moon dataset, and (iv) comparison with other approximate inference approaches on nine UCI regression datasets \cite{Dua2019uci}. The KBNN implementation merely requires matrix operations and is realized in PyTorch. The performance of the methods is assessed by means of the root mean square error (RMSE) for regression tasks, the accuracy for classification tasks, the negative log-likelihood (NLL\arXiv{, explained in the supplementary material}) for quantifying the uncertainty, and the training time. A PC with Intel i7-8850H CPU, 16GB RAM but without GPU is used.

\begin{figure}[tb]
  \begin{center}
    \includegraphics[width=0.45\textwidth]{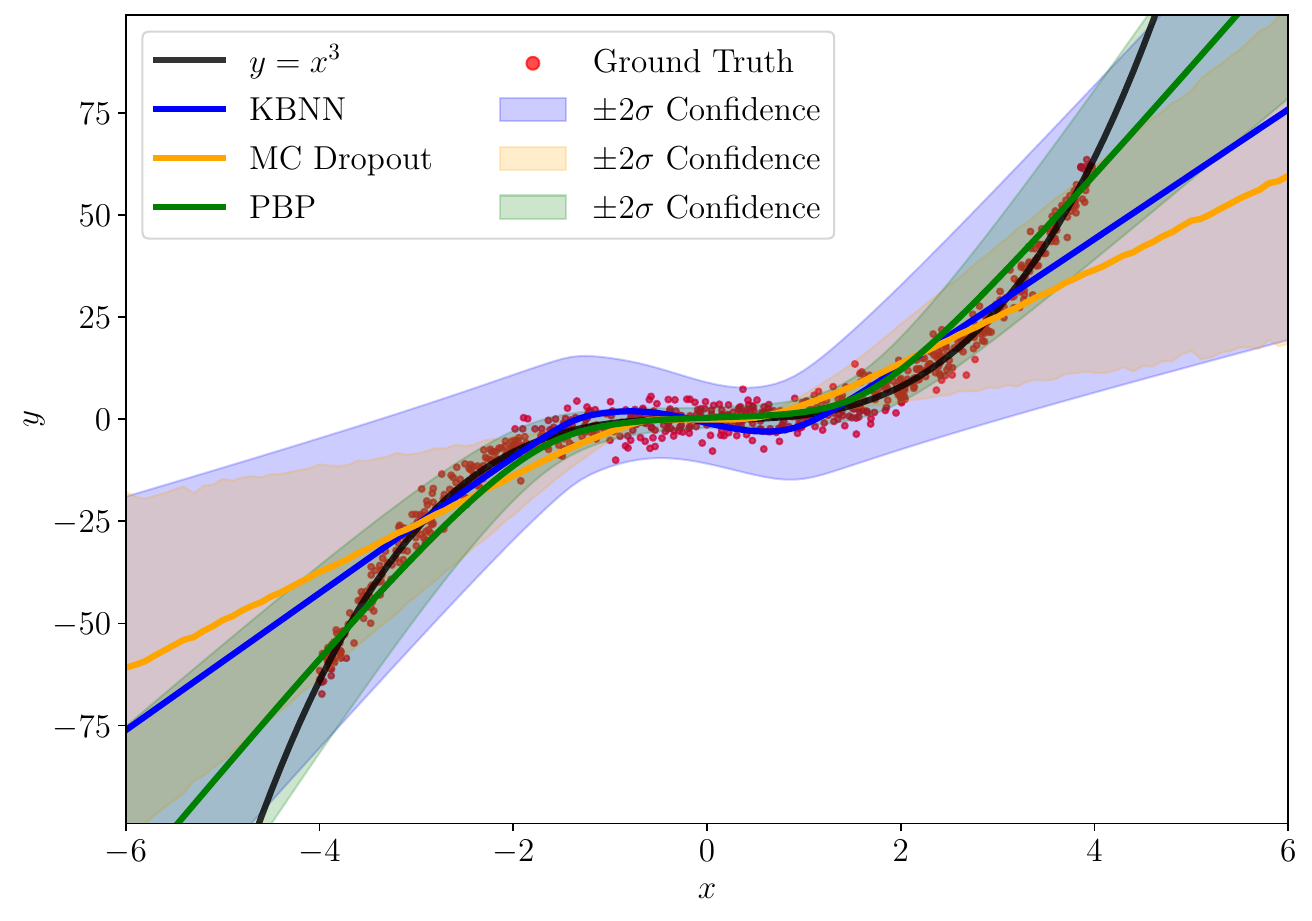}
  \end{center}
	\vspace{-2mm}
  \caption{Predictions of KBNN, MC Dropout and PBP trained for one epoch on the regression task $y = x^3 + \epsilon_n$.} 
  \label{fig:toy_regression}
\end{figure}

\paragraph{Regression on Synthetic Data}
\label{sec:experiments_regression-synth}



We generate a synthetic dataset with $800$ data instances from the polynomial $ y = x^{3} + \epsilon_{n} $, where $ \epsilon_{n}\sim\SN(0, 9) $ similar to \citet{hernandez2015probabilistic}, while $x$ is sampled uniformly from the interval $\left[-4, 4\right]$. We use a standard MLP with one hidden layer and 100 hidden neurons, and ReLU activation for the hidden layer. The output activation is linear. We compare KBNN with PBP \cite{hernandez2015probabilistic} and Monte Carlo (MC) Dropout \cite{gal16dropout}. For both PBP and MC Dropout we use the implementations of the authors
. For MC Dropout we use dropout probability 0.1, same as the authors used for regression tasks \cite{gal16dropout}. All methods merely use one epoch for training in order to simulate an online learning scenario. In Fig.~\ref{fig:toy_regression} the results of all methods are depicted. KBNN, PBP and MC Dropout are able to approximate the underlying nonlinear function and perform similar, where PBP and MC Dropout tend to underestimate the aleatoric uncertainty. 

\begin{figure*}[t]
    \centering
    \includegraphics[width=0.91\textwidth]{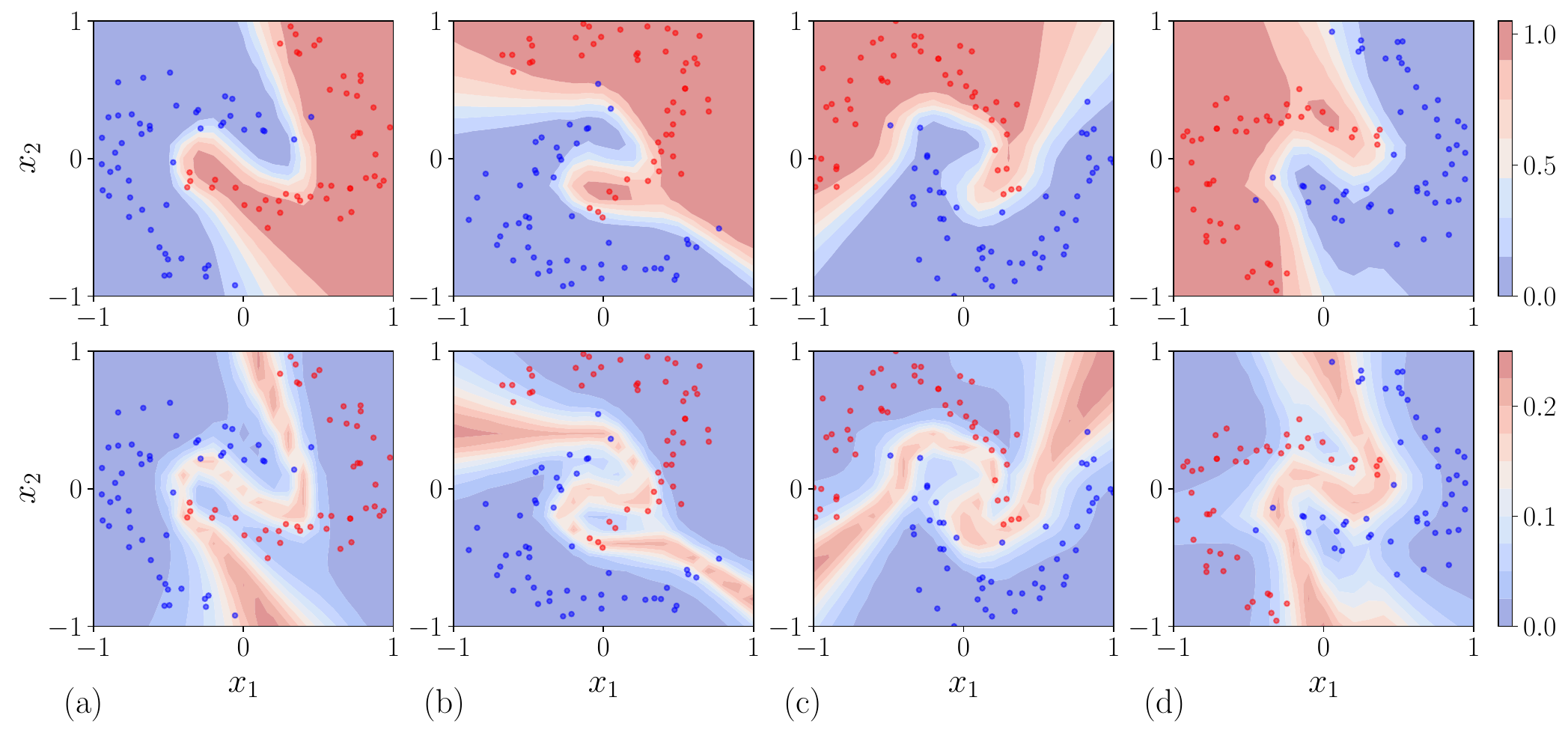}
    \caption{Online learning on the non-stationary Moon dataset. Predictive mean (top row) and variance (bottom row) after rotating by (a) 60, (b) 120, (c) 180, and (d) 240 degrees, respectively.}
    \label{fig:online_moon}
\end{figure*}


\paragraph{Binary Classification}\label{subsec:classification}
To validate the capabilities of the KBNN for sequential/online learning in binary classification tasks, we perform classification on the Moon dataset \cite{pedregosa2011scikit} with 1,500 instances in total. The data is presented in a continuous stream to the learning algorithm, starting with a single data instance. Training ends, when $90\%$ of the dataset, i.e., 1,350 instances are processed. We measure the performance of the model on the remaining $10\%$ of the instances during training to evaluate the learning progress. 
To demonstrate learning for multiple layers we use two hidden layers, each with 10 neurons. The hidden activations are ReLU, the output activation is a sigmoid function. 

\begin{table}[t]
\centering
\caption{Sequential learning on the Moon dataset.}
\vspace{-2mm}
\label{table:moon}
\tabcolsep=0.12cm
\scalebox{0.8}{\begin{tabular}{lccc}
\hline
$\#$ data & Accuracy & NLL & Training Time / s \\
\hline
$5$ & $47.53\%\pm 0.60\%$ & $0.18\pm 0.02$ & $0.01\pm 4.00*10^{-4}$ \\
$50$ & $88\%\pm 1.63\%$ & $0.11\pm 0.01$ & $0.13\pm 1.20*10^{-3}$ \\
$500$ & $92.07\%\pm 2.28\%$ & $0.05\pm 0.01$ & $1.25\pm 3.53*10^{-3}$ \\
$1,000$ & $97.87\%\pm 2.33\%$ & $0.03\pm 3.60*10^{-3}$ & $2.49\pm 8.13*10^{-3}$ \\
$1,350$ & $99.93\%\pm 0.20\%$ & $0.03\pm 3.43*10^{-3}$  & $3.40 \pm 5.72*10^{-3}$\\
\hline
\end{tabular}}
\end{table}

Table~\ref{table:moon} lists how the accuracy and NLL of the KBNN on the test set evolve for an increasing number of processed training data. These results are averages over ten random trials. After several seconds of training the proposed model achieves a high accuracy and low NLL on the test set. \arXiv{Fig.~\ref{fig:classification_moon} in the supplementary material depicts the learned decision boundaries (top) and the predictive variance (bottom). 
The model predicts higher uncertainties along the decision boundary, where both classes intermix, which is as expected.}

\paragraph{Online Learning}
\label{sec:experiments_online}
In order to validate the online learning capability of the KBNN on non-stationary data, it is applied to the classification of a rotating Moon dataset, similar as in \cite{kurle2019continual}. We use the same network architecture as in Sec.~\ref{subsec:classification}. At the first iteration, we train the KBNN with the standard Moon dataset comprising 1,500 instances. Then, we continue training the KBNN for 18 iterations, where for each iteration the dataset is rotated by 20 degrees and comprises only 100 data instances. 
Fig.~\ref{fig:online_moon} shows the changing decision boundary (predictive mean) after 60, 120, 180, and 240 degrees rotations. After each rotation, the KBNN can always efficiently adapt to the new data distributions.

\begin{table*}[t]
\small
\centering
\caption{RMSE on the test set for nine different UCI regression datasets.}\vspace{-2mm}
\tabcolsep=0.15cm
\scalebox{0.75}{
\begin{tabular}{l | cc | ccccc}
\hline
Dataset & $\mathit{N}$ & $\mathit{d}$ & SVI & MCMC & PBP & KBNN 1  & KBNN 10  \\\hline
Boston & $506$ & $13$ & $3.434\pm 0.131$ & $ \mathbf{2.553\pm 0.027}$ & $2.740\pm 0.095$ & $3.893\pm 0.200$ & $2.695\pm 0.155$ \\
Concrete & $1,030$ & $8$ & $7.597\pm 0.283$ & $6.227\pm 0.108$ & $5.874\pm 0.054$ & $8.396\pm 0.497$ & $\mathbf{5.703\pm 0.183}$ \\
Energy & $768$ & $8$& $4.025\pm 0.074$ & $\mathbf{0.906\pm 0.049}$ & $3.274\pm 0.049$ & $4.155\pm 0.087$ & $2.404\pm 0.259$ \\
Wine & $4,898$ & $11$& $0.726\pm 0.007$ & $\mathbf{0.656\pm 0.004}$ & $0.667\pm 0.002$ & $0.719\pm 0.011$ & $0.666\pm 0.006$ \\
Naval & $11,934$ & $16$ & $0.025\pm 0.012$ & $0.008\pm 0.001$ & $0.006\pm 6.12*10^{-5}$ & $0.034\pm 0.005$ & $\mathbf{0.004\pm 0.001}$ \\
Yacht & $308$ & $6$ & $1.157\pm 0.222$ & $0.879\pm 0.294$ & $\mathbf{0.867\pm 0.047}$ & $3.752\pm 0.240$ & $1.584\pm 0.178$ \\
Kin8nm  & $8,192$ & $8$ & $0.101\pm0.002$ &  $\mathbf{0.081\pm0.003}$ & $0.100\pm 0.003$ & $ 0.174\pm 0.006$ & $0.110\pm 0.005$ \\
Power  & $9,568$ & $4$ & $4.419\pm0.046$ & $287.227\pm200.167$  & $\mathbf{4.060\pm 0.009}$ & $4.243\pm 0.011$ & $4.167\pm 0.034$ \\
Year  & $515,345$ & $90$ & $25.163\pm1.990$ & $\mathbf{NA}$ & $8.879\pm 0.004$ & $8.887\pm 0.014$ & $\mathbf{8.874\pm 0.015}$ \\
\hline
\end{tabular}}
\label{table:UCI-performance-rmse}
\end{table*}

\begin{table*}[t]
\centering
\caption{NLL on the test set and training time on the training set for the UCI regression datasets.}
\vspace*{-2mm}
\tabcolsep=0.09cm
\scalebox{0.76}{
\begin{tabular}{l | ccccc | ccccc}
\hline
 &\multicolumn{5}{c}{NLL} & \multicolumn{5}{c}{training time} \\\hline
Dataset & SVI & MCMC & PBP & KBNN 1 & KBNN 10 & SVI & MCMC & PBP & KBNN 1 & KBNN 10\\\hline
Boston & $2.690\pm 0.041$ & $\mathbf{2.346\pm 0.010}$ & $2.421\pm 0.037$ & $3.183\pm 0.046$ & $2.767\pm 0.392$ 
& $21.4$ & $446.0$ & $8.2$ & $\mathbf{0.8}$ & $8.7$\\
Concrete & $3.446\pm 0.042$ & $3.236\pm 0.024$ & $\mathbf{3.119\pm 0.010}$ & $3.676\pm 0.119$ & $8.428\pm 0.946$ 
& $22.5$ & $481.7$ & $12.7$ & $\mathbf{1.7}$ & $17.5$\\
Energy & $2.877\pm 0.026$ & $\mathbf{1.315\pm 0.179}$ & $2.680\pm 0.020$ & $3.078\pm 0.015$ & $2.394\pm 0.159$ 
& $21.6$ & $405.9$ & $10.2$ & $\mathbf{1.2}$ & $13.2$\\
Wine & $1.107\pm 0.011$ & $\mathbf{1.003\pm 0.006}$ & $1.014\pm 0.002$ & $1.529\pm 0.202$ & $1.127\pm 0.122$ 
& $23.8$ & $520.3$ & $49.1$ & $\mathbf{8.3}$ & $86.7$\\
Naval & $-1.817\pm 0.179$ & $-3.424\pm 0.091$ & $\mathbf{-3.736\pm 0.021}$ & $1.266\pm 0.255$ & $0.128\pm 0.008$
& $42.8$ & $367.0$ & $116.1$ & $\mathbf{20.5}$ & $205.3$\\
Yacht & $\mathbf{1.435\pm 0.336}$ & $1.649\pm 0.457$ & $1.558\pm 0.036$ & $3.033\pm 0.022$ & $2.325\pm 0.055$
& $21.5$ & $357.4$ & $5.9$ & $\mathbf{0.5}$ & $5.0$\\
Kin8nm & $-0.869\pm 0.026$ & $\mathbf{-1.094 \pm 0.034}$ & $-0.882\pm 0.026$ & $  -0.255\pm 0.139 $ & $ -0.758\pm 0.043$ 
& $57.6$ & $1200.0$ & $107.9$ & $\mathbf{20.2}$ & $204.0$ \\
Power & $3.162\pm 0.071$ & $598.460 \pm 422.969$ & $ \mathbf{2.820 \pm 0.002}$ & $ 3.062 \pm 0.040 $ & $ 2.922\pm 0.015$ 
& $56.6$ & $769.9$ & $93.4$ & $\mathbf{20.0}$ & $208.6$ \\
Year & $6.801 \pm 0.765$ & $\mathbf{NA}$ & $ \mathbf{3.588\pm 0.001}$ & $ 4.638 \pm 0.219 $ & $ 4.315\pm 0.116$ 
& $5419.7$ & $\mathbf{NA}$ & $5694.9$ & $\mathbf{2021.7}$ & $20801.8$ \\
\hline
\end{tabular}}
\label{table:UCI-performance-nll}
\end{table*}

\paragraph{UCI Regression Datasets}
\label{subsec:UCI-regression}
In this section we compare the KBNN to SVI, MCMC and PBP for regression tasks on nine UCI datasets: Boston, Concrete, Energy, Wine, Naval, Yacht, Kni8nm, Power and Year. These datasets are commonly used for BNN performance evaluation (cf. \citet{hernandez2015probabilistic}). Like in the previous sections, the datasets are split into random train and test sets with $90\%$ and $10\%$ of the data, respectively. For SVI and MCMC we use implementations build in the probabilistic programming package Pyro \cite{bingham2018pyro}
. All methods are compared using the same network architecture with one hidden layer comprising 50 units and ReLU activations as proposed in \citet{hernandez2015probabilistic}. We use $40$ epochs for PBP as in \citet{hernandez2015probabilistic} and $5,000$ epochs for SVI, after which the trained models converged well mostly. MCMC is conducted with NUTS and we draw $100$ samples from the posterior. Although KBNN is designed for online learning, i.e., processing the data only once, we also executed KBNN with 10 epochs of training---denoted KBNN~10 in the following---to improve the performance. We repeat the experiments ten times with random initializations and average the results.

Tables~\ref{table:UCI-performance-rmse} and \ref{table:UCI-performance-nll} show the average RMSE and NLL on the test sets as well as the training time. 
KBNN~1, i.e., the online version, achieves a performance being close to other methods on some datasets while requiring significantly less training time for all datasets. Particularly compared to SVI, the performance gap between KBNN~1 and SVI is narrow. 
KBNN~10 outperforms SVI in most cases and PBP and MCMC on some datasets. For the Concrete, Naval and Year datasets, KBNN even outperforms MCMC and PBP in terms of RMSE and PBP also on Boston and Energy. 
For an increasing number of epochs the NLL value of the KBNN increases in-between for the Boston, Concrete and Wine datasets. \arXiv{A plot of the NLL against the number of epochs showing this behavior can be found in the supplementary material.} This increase is caused by too low variances.

KBNN~1 is clearly faster than the other methods. The training time roughly grows linearly with the number of data instances. Thus, compared to SVI, which is designed for scaling well with large datasets \cite{Zhang2019}, KBNN has a runtime advantage on smaller datasets while this gap closes for larger datasets and more epochs of training. However, it is worth mentioning that as a method with online learning capabilities in contrast to SVI, MCMC and PBP, our method shows great single sample learning efficiency. 
If SVI or PBP learn for only one epoch, their performance significantly drops and is worse than KBNN~1, especially for the small datasets. 
Averaged over all datasets, the time of KBNN~1 to process a single input is $1.659 \pm 0.041$ ms, which is promising for real-time tasks.

We also performed a series of experiments with either a different number of hidden neurons or different number of hidden layers to assess the scalability of the KBNN. \arXiv{For details please be referred to the supplementary material.}

\section{Discussion}
\label{sec:discussion}
In this paper we introduced an approach to perform sequential and online learning of BNNs via assumed Gaussian filtering/smoothing. The state of the art in training BNNs are VI-based methods. Although being Bayesian, these methods treat training as an optimization problem. Instead, the proposed KBNN approach is fully Bayesian in the sense that the training strictly aims at (approximately) solving Bayes' rule~\eqref{eq:bayes}. Utilizing concepts from Bayesian filtering and smoothing allows updating the mean and covariance of the weight posterior in closed form and in an online fashion, which are two key features compared to the state of the art. 


\paragraph{Strengths}
Given the Assumptions~1 and 2, which do not hinder the learning abilities of the KBNN in practice, our approach performs approximate but fully Bayesian inference for training. For ReLU activations it provides moment matching Gaussian approximations of the predictive and posterior distribution. 
This is clearly an advantage compared to other methods that rely on stochastic gradient descent. The absence of gradients proves to be data efficient and enables the usage of activation functions that cannot be used in gradient-based learning, e.g., the Heaviside activation or non-differentiable activation schemes.

A second advantage of the proposed method is the ability of learning from sequential data streams without retraining. As shown in the conducted experiments every data instance has to be seen only once during training while still achieving decent performance on the respective test set. This can be especially useful in online learning scenarios or in the context of model-based reinforcement learning where retraining is needed to update the model of the environment.

The update rule of the weights' means \eqref{eq:muwupdate} can more abstractly be written as $\text{new} = \text{old} + \L\cdot \Delta$, which is similar to the backpropagation update rule. But instead of a scalar learning rate being a hyper-parameter, KBNN uses the matrix $\L$, i.e., it uses a matrix-valued, intrinsically calculated learning rate where each weight obtains its individual rate.


\paragraph{Limitations}
To keep the probability distribution of the network parameters manageable in complexity, independence between the weights of different neurons is assumed (cf. Assumption~1). Abandoning this independence would require the calculation of cross-covariances between neurons. This affects our approach mainly in two ways. First, the memory and computational demand for additionally calculating these terms increases quadratically with the number of neurons per layer. 
Second, the necessary additional calculation of $\E[f(a_i)\cdot f(a_j)]$ to obtain the cross-covariance between the activations of neurons $i,j = 1\ldots M_l$ in the forward pass is challenging. It is questionable if an analytic solution even for ReLU activations exists \cite{wu2019}.

The considered assumptions significantly simplify the necessary computations and enable closed-form calculations of the quantities of interest. While Assumption~2 is very reasonable for regression tasks, it is not well justified for classification tasks where one would rather want to use for instance a Bernoulli distribution for the output \cite{hennig2020bayesian}. The use of distributions other than a Gaussian as in our case would only be possible if the Kalman filter in the last layer is replaced by more advanced filters such as a particle filter, which uses sampling to approximate the posterior \cite{saerkkae_2013}. The Gaussian assumption seems not to impair the performance of the KBNN in classification tasks, at least in the conducted experiments. 

\paragraph{Open Issues and Future Work}
For multi-class classification problems it is common to use a soft-max activation at the output layer. Unfortunately, there generally is no closed-form solution of \eqref{eq:muz} and \eqref{eq:varz} if $f(.)$ is a soft-max function. At least \eqref{eq:muz} can be calculated if the mean-field approximation is applied \cite{Lu2021}. Using a hard-max activation instead of soft-max allows a closed-form solution. 

PBP learns its hyper-parameters, which is not the case for the KBNN. To avoid tedious hyper-parameter tuning, adopting a hierarchical Bayesian approach as in \cite{hernandez2015probabilistic} is part of future work. 

Convolutions and pooling are linear and piece-wise linear operations, respectively. Thus, future work is also devoted to extend the KBNN for processing image data.


\section*{Acknowledgements}
This work was partially supported by the Baden-W\"urttemberg Ministry of Economic Affairs, Labor, and Tourism within the KI-Fortschrittszentrum ``Lernende Systeme and Kognitive Robotik'' under Grant No. 036-140100.


\newpage
\appendix
\onecolumn

\arXiv{\input{sm}}

%

\end{document}

%% file: figures/PGM/PGM_tikz2.tex
\begin{tikzpicture}

\Vertex[label=$z^{l+1}_n$, x=4]{y}
\Vertex[label=$a_n^{l}$, x=2]{al}
\Vertex[x=2.1, y=-0.3, opacity=0, size=0.1, style={color=white, draw opacity=0}]{helpar}
\Vertex[x=1.9, y=-0.3, opacity=0, size=0.1, style={color=white, draw opacity=0}]{helpal}
\Vertex[label=$\vw_n^l$, x=1, y=1]{Wlm}
\Vertex[y=-0.45, opacity=0, size=0.1, style={color=white}]{help}
\Vertex[label=$\vz^{l}$]{zl}
\Vertex[x=-0.9, opacity=0, size=0.1, style={color=white}, label=$\mathbf{\cdots} \quad $]{help2}
\Vertex[x=5, opacity=0, size=0.1, style={color=white}, label=$\quad \mathbf{\cdots}$]{help3}

\EdgesNotInBG
\Edge[Direct](al)(y) 
\Edge[path={y,{4,-1},{2.1,-1}}, color=orange, label=smoothing, Direct, style={rounded corners=.1cm}](y)(helpar)
\Text[x=3.05,y=-0.85, fontsize=\scriptsize, color=orange]{smoothing I}

\Edge[Direct](Wlm)(al) 

\Edge[path={helpal,{1.9,-1},{0,-1}}, color=orange, label=smoothing, Direct, style={rounded corners=.1cm}](helpal)(help)
\Text[x=0.9,y=-0.85, fontsize=\scriptsize, color=orange]{smoothing II}

\Edge[Direct](zl)(al) 
\Edge[Direct](help2)(zl)
\Edge[Direct](y)(help3)
\Plane[x=-.5,y=-.5,width=2.5,height=1, NoFill, style={color=orange, line width=0.3mm, rotate=45, rounded corners=.4cm}]{Plane}
\Plane[x=-0.5,y=0.5,width=2.0,height=4, style={line width=0.3mm, rounded corners=.1cm}, InBG=true]{Plane}

\Text[x=3.8,y=1.3, fontsize=\scriptsize]{Neuron $n$}
\end{tikzpicture}

%% file: sm.tex
\section*{Supplementary Material}
\label{sm:Supplementary}

In the supplementary material the following additional derivations and experiments can be found:
\begin{enumerate}[label=\textbf{Section \Alph*}, leftmargin =*, wide=0pt]
    \item Closed-form calculations/approximations of expected values depending on the NN's activation functions. These quantities are necessary in both the forward pass and the backward pass of the KBNN.
    \item Derivation of the cross-covariance $\C_{wza}$, which is used in \eqref{eq:cwza} being part of the backward pass.
    \item Contains the definition of the negative log-likelihood metric used in the experiments for quantifying the predictive performance of the various methods examined. 
    \item Contains an additional evaluation of the KBNN on the Moon and the Circles datasets. Further, we show how the uncertainty of a trained KBNN for binary classification evolves in areas being far away from the training data.
    \item Results on the synthetic regression task of Sec.~\ref{sec:experiments_regression-synth} for a significantly lower number of training data instances.
    \item Shows how the RMSE and the NLL evolve on the considered UCI datasets with an increasing number of training epochs. 
    \item Experiments showing the influence of different numbers of hidden layers and different numbers of hidden neurons on KBNN's performance on the UCI datasets.
\end{enumerate}

\setcounter{section}{0}
\renewcommand{\thesection}{\Alph{section}}

\section{Expected Values for Common Activation Functions}
\label{sm:exp}
The mean \eqref{eq:muz} and variance \eqref{eq:varz} required for the forward pass and the covariance $(\sigma_{az}^n)^2$ needed for the Kalman gain $\vec k_n$ in \eqref{eq:smootheda} for the backward pass depend on the used activation function $f(.)$. In the following, these quantities are given for piece-wise linear activations in Sec.~\ref{sm:exp_linear} and for the sigmoid activation function in Sec.~\ref{sm:exp_sigmoid}. The results are taken from \citet{huber2020bayesian} and we refer to their work for detailed derivations. 

\subsection{Piece-wise Linear}
\label{sm:exp_linear}
A general piece-wise activation is given by	$f(a) = \max(\alpha \cdot a, \beta\cdot a)$ with $\alpha \in [0,1]$, $\beta \ge 0$, and $\alpha \le \beta$ which includes ReLU as a special case for $\alpha = 0$, $\beta = 1$. The mean value \eqref{eq:muz} is given by
\begin{equation*}
    \mu_z^{l+1,n} 
	= \E[f(a_n)] 
	= \alpha \cdot \mu_a^n +\, (\beta - \alpha) \cdot \left( \mu_a^n \cdot\, \phi\left(\tfrac{\mu_a^n}{\sigma_a^n}\right) + p_a\right),\quad~
\end{equation*}
with the \emph{probit} function $\phi(a) = \nicefrac{1}{2}\cdot(1+\operatorname{erf}(\nicefrac{a}{\sqrt{2}}))$ containing the Gaussian error function $\operatorname{erf}(.)$ and $p_a\DEF(\sigma_a^n)^2\cdot \SN(0\,|\, \mu_a^n, (\sigma_a^n)^2)$. 

The variance \eqref{eq:varz} is given by
\begin{equation*}
	(\sigma_z^{l+1,n})^2
	= \E\left[f(a_n)^2\right] - (\mu_z^{l+1,n})^2 
	= \,\alpha^2\cdot\gamma +\, c\cdot\Big(\gamma \cdot\,\phi\left(\tfrac{\mu_a^n}{\sigma_a^n}\right) + \mu_a^n\cdot p_a\Big) - (\mu_z^{l+1,n})^2~,
\end{equation*}
with $c \DEF \left(\beta^2-\alpha^2\right)$ and $\gamma \DEF (\mu_a^{n})^2 + (\sigma_a^n)^2$.

The covariance $(\sigma_{za}^n)^2$ is given by
\begin{align*}
    (\sigma_{za}^n)^2 = \E [a_n \cdot f(a_n)] - \mu_a^{n} \cdot \mu_z^{l+1,n}.
\end{align*}
For piece-wise linear activations this expectation value can be calculated exactly resulting in
\begin{align*}
    (\sigma_{za}^n)^2 =& \, \alpha\cdot \gamma +\, (\beta-\alpha)\cdot\left(\gamma\cdot\,\phi\Big(\tfrac{\mu_a^n}{\sigma_a^n}\right) + 
     \mu_a^n\cdot p_a\Big) - \mu_a^{n} \cdot \mu_z^{l+1,n}.
\end{align*}

\subsection{Sigmoid}
\label{sm:exp_sigmoid}
The sigmoid activation function is defined as $f(a_n) = s(a_n) \DEF \frac{1}{1 + \text{e}^{-a_n}}$. In contrast to the piece-wise linear activation discussed above, the sigmoid allows no closed-form calculation of the required quantities. However, they can be approximated closely in closed form when replacing the sigmoid by the probit function. For this purpose we use $s(a_n) \approx \phi(\lambda \cdot a_n)$ with $\lambda \DEF \sqrt{\nicefrac{\pi}{8}}$ \cite{murphy2012machine}. Using this relation the mean $\eqref{eq:muz}$ can be approximated via
\begin{equation*}
	\mu_z^{l+1,n} = \E[s(a_n)] \approx \phi\hspace{-.5mm}\left(\tfrac{\lambda\cdot\mu_a^n}{t_n}\right),
\end{equation*}
with $t_n \DEF \sqrt{1+\lambda^2\cdot (\sigma_a^{n})^2}$\,. 

The variance \eqref{eq:varz} is given by
\begin{equation*}
    (\sigma_z^{l+1,n})^2 =  \E\left[ s(a_n)^2\right] - \mu_z^{l+1,n} 
    \approx \mu_z^{l+1,n}\cdot(1-\mu_z^{l+1,n})\cdot(1 - \tfrac{1}{t_n})~,
\end{equation*}
using the same approximation.

For the covariance the usage of the probit function yields the tight approximation
\begin{align*}
    (\sigma_{za}^n)^2 \approx \tfrac{\lambda \cdot (\sigma_a^n)^2}{t_n} \cdot \SN \left( \tfrac{\lambda \cdot \mu_a^n}{t_n}\,\big|\, 0, 1 \right).
\end{align*}

These formulae of the mean, variance, and covariance can be straightforwardly applied to hyperbolic tangent activations, as sigmoid and hyperbolic tangent are related by means of the linear transformation $\tanh(a) = 2\cdot s(a) + 1$. 

\section{Derivation of the Cross-Covariance $\C_{wza}$}
\label{sm:Cwza}
For an arbitrary layer the cross-covariance $\C_{wza}$ in \eqref{eq:cwza} is defined as the expected value
\begin{align*}
    \C_{wza} 
    &= \E \left[ \left(\begin{bmatrix} \vw \\ \vz \end{bmatrix} - \begin{bmatrix} \vmu_w \\ \vmu_z^- \end{bmatrix}\right) \cdot (\va - \vmu_a^- )\T \right] 
\end{align*}
with $ \C_{wza} \in \NewR^{M_{l}\cdot(M_{l-1}+1+M_{l-1}) \times M_l}$. This matrix contains two types of entries, i.e.,
\begin{align*}
    \E_1 &= \E [ \vz \cdot (\vw_i\T \cdot \vz) ] 
    = \C_{z}^- \cdot \vmu_w^{i} + \vmu_{z}^- \cdot ((\vmu_w^{i})\T \cdot \vmu_z^-)~, \\
    \E_2 &= \E [ \vw_i \cdot (\vw_j\T \cdot \vz) ]  
    = \sum_n \E [ \vw_i \cdot w_{j,n} ] \cdot \E [ z_n ]
    = \begin{cases} \C_w^{i} \cdot \vmu_z^- + \vmu_w^{i} \cdot ((\vmu_w^{i})\T \cdot \vmu_z^- ) & \text{for } i = j \\ 
    \vmu_w^{i} \cdot ( (\vmu_w^{i})\T \cdot \vmu_z^- ) & \text{for } i \neq j \end{cases} 
\end{align*}
for $i,j = 1\ldots M_l$. The terms without covariances cancel out and we obtain
\begin{align*}
\C_{wza}&= \begin{bmatrix}
                        \C_{w}^{1} \cdot \vmu_z^{l,-} & \cdots & 0 \\
                        \vdots & \ddots & \vdots \\
                        0 & \cdots & \C_{w}^{M_l} \cdot \vmu_z^{l,-} \\
                        \C_z^{l,-} \cdot \vmu_{w}^{1} & \cdots & \C_z^{l,-} \cdot  \vmu_{w}^{M_l}
                    \end{bmatrix}       \nonumber 
                = \begin{bmatrix} \diag \left( \C_{w}^{1} \cdot \vmu_z^{l,-}, \ldots , \C_{w}^{M_l} \cdot \vmu_z^{l,-} \right) \\
        \C_z^{l,-}  \cdot\vmu_{w}^{1}\, \cdots\, \quad \C_z^{l,-} \cdot \vmu_{w}^{M_l}
    \end{bmatrix}~. 
\end{align*}

\begin{figure*}[t]
    \centering
    \includegraphics[width=1\textwidth]{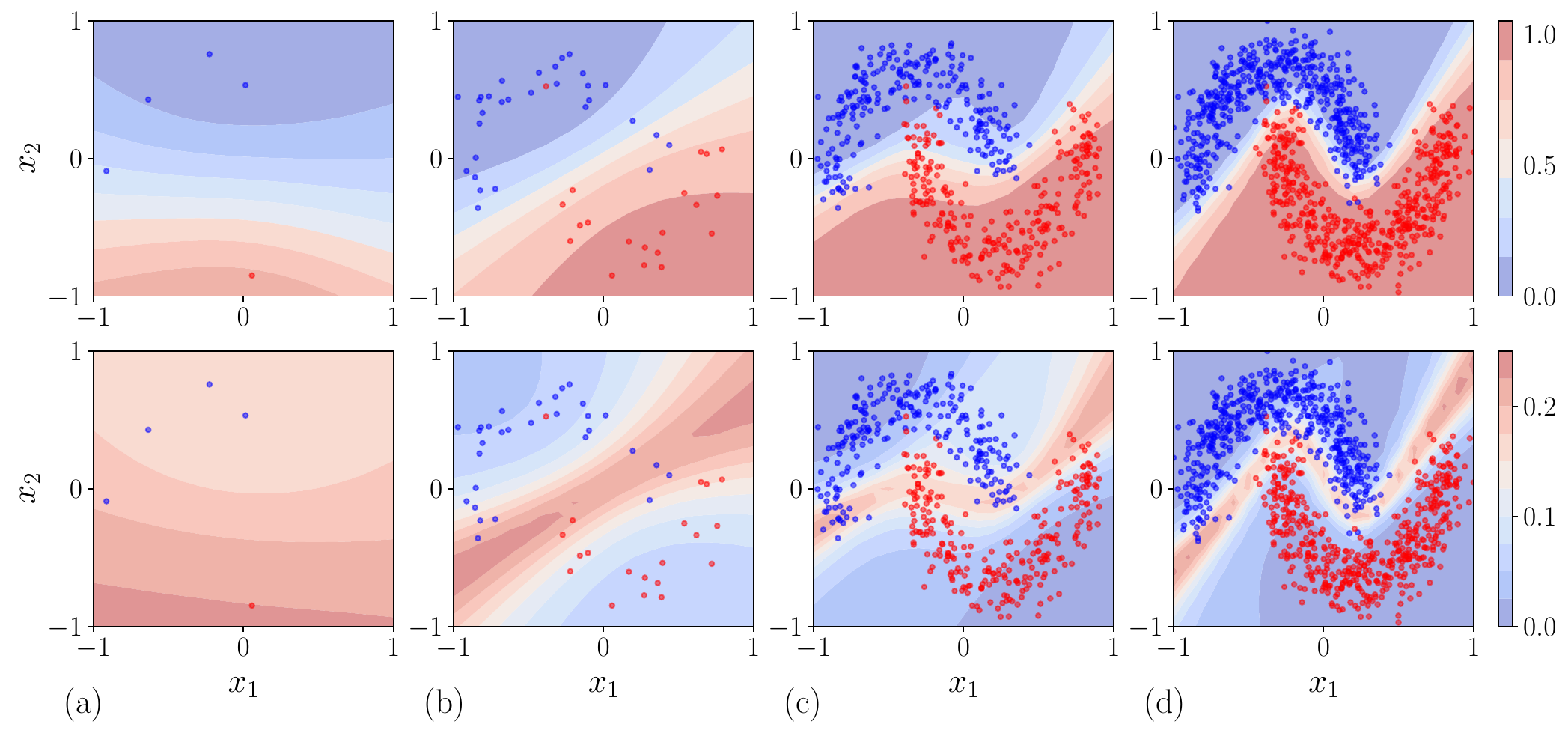}
    \vspace*{-5mm}\caption{Sequential learning of the predicted mean (first row) and the variance of predictions (second row) on the moon dataset, for (a) five, (b) 50, (c) 500 and (d) 1,000 data instances. The samples of class 1 are drawn as blue dots, while the samples of class 2 are drawn as red dots. With an increasing number of data instances, the learnt decision boundary becomes increasingly sharp and accurate.}
    \label{fig:classification_moon}
\end{figure*}
\begin{figure*}[t]
    \centering
    \includegraphics[width=1\textwidth]{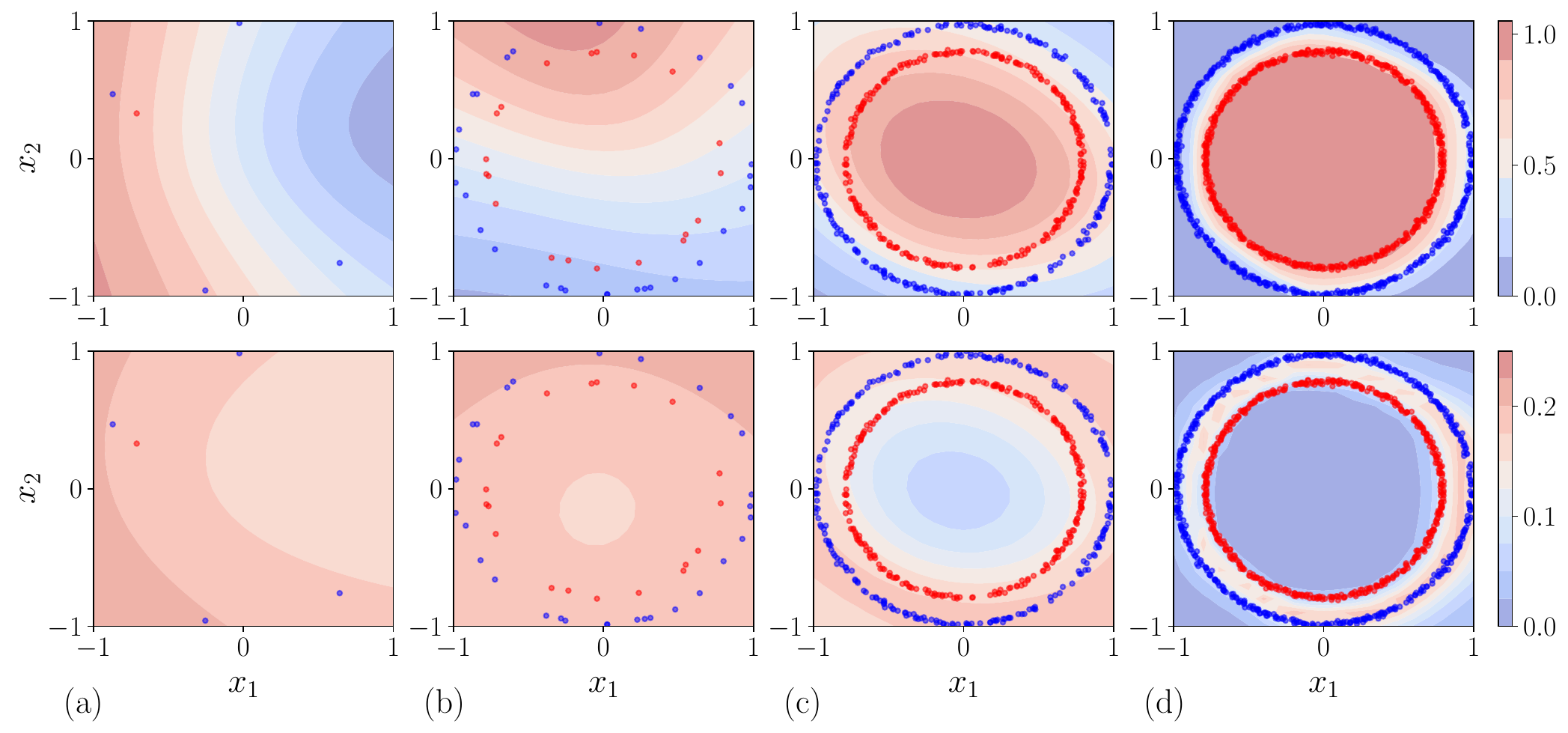}
    \vspace*{-5mm}\caption{Sequential learning of the predicted mean (first row) and the variance of predictions (second row) on the circles dataset, for (a) five, (b) 50, (c) 500 and (d) 1,000 data instances. The samples of class 1 are drawn as blue dots, while the samples of class 2 are drawn as red dots. With an increasing number of data instances, the learnt decision boundary becomes increasingly sharp and accurate.}
    \label{fig:classification_circles}
\end{figure*}

\section{Definition of the Negative Log-Likelihood}
\label{sm:nll}
To evaluate model uncertainties in Sec.~\ref{sec:experiments}
, we use the average negative log-likelihood (NLL) on test sets as a metric. In the following let $\mu(\vx)$ and $\sigma(\vx)$ be functions that are set to the calculated predictive mean and standard deviation for a given input $\vx$ by the respective examined method. In case of the KBNN these functions correspond to \eqref{eq:muz} and \eqref{eq:varz}, respectively, for $l=L$. For a data set with test data instances $(\vx_i, \vy_i)$, $i=1\ldots N$, the average NLL is defined as
\begin{align}
    \boldsymbol\ell_\mathrm{NLL} &= -\frac{1}{N}\sum_{i=1}^{N}\textup{log} \ \SN(\vy_i|\mu(\vx_i), \sigma(\vx_i)) \nonumber \\
    &= \frac{1}{2N}\sum_{i=1}^{N}\left ( \frac{(\vy_i - \mu(\vx_i))^2}{\sigma(\vx_i)^2}  + \log \sigma(\vx_i)^2 \right ) +\tfrac{1}{2}\log 2\pi 
    \label{eq:nll}
\end{align}
assuming that $\vy\sim \mathcal{N}(\mu(\vx), \sigma(\vx)^2)$. It can be seen that the first term of the sum in \eqref{eq:nll} penalizes deviations of the predicted mean $\mu(\vx_i)$ from the ground truth $\vy_i$ with at the same time small predictive variance $\sigma(\vx_i)^2$. The second term of the sum encourages lower uncertainties.

\section{Additional Experiments for Binary Classification}
\label{sm:binary}

\subsection{Learning Process on the Moon and Circles Dataset}
To demonstrate the sequential/online learning progress of the proposed KBNN, classification tasks on the Moon and the Circles datasets \cite{pedregosa2011scikit} are conducted. 
The experimental setup, i.e., the network architecture and the provision of the training data, is as described in Sec.~\ref{subsec:classification}. The progress on sequential learning in terms of the predictive mean and variance are depicted in Fig.~\ref{fig:classification_moon} for the Moon dataset and in Fig.~\ref{fig:classification_circles} for the Circles dataset. It can be seen that the initially random decision boundary continuously improves with an increasing number of data instances.

\subsection{Investigation of the Uncertainty for Unknown Data Regimes}

In the following we investigate the uncertainty quantification for binary classification problems. For regression problems, the experiment in Sec.~\ref{sec:experiments_regression-synth} 
shows that the uncertainty grows when moving away from the training data. In binary classification, however, the network's output is calculated by means of the sigmoid activation function, for which the Gaussian approximation of the output can be inappropriate. 

\begin{figure*}[t]
    \centering
    \includegraphics[width=0.6\textwidth]{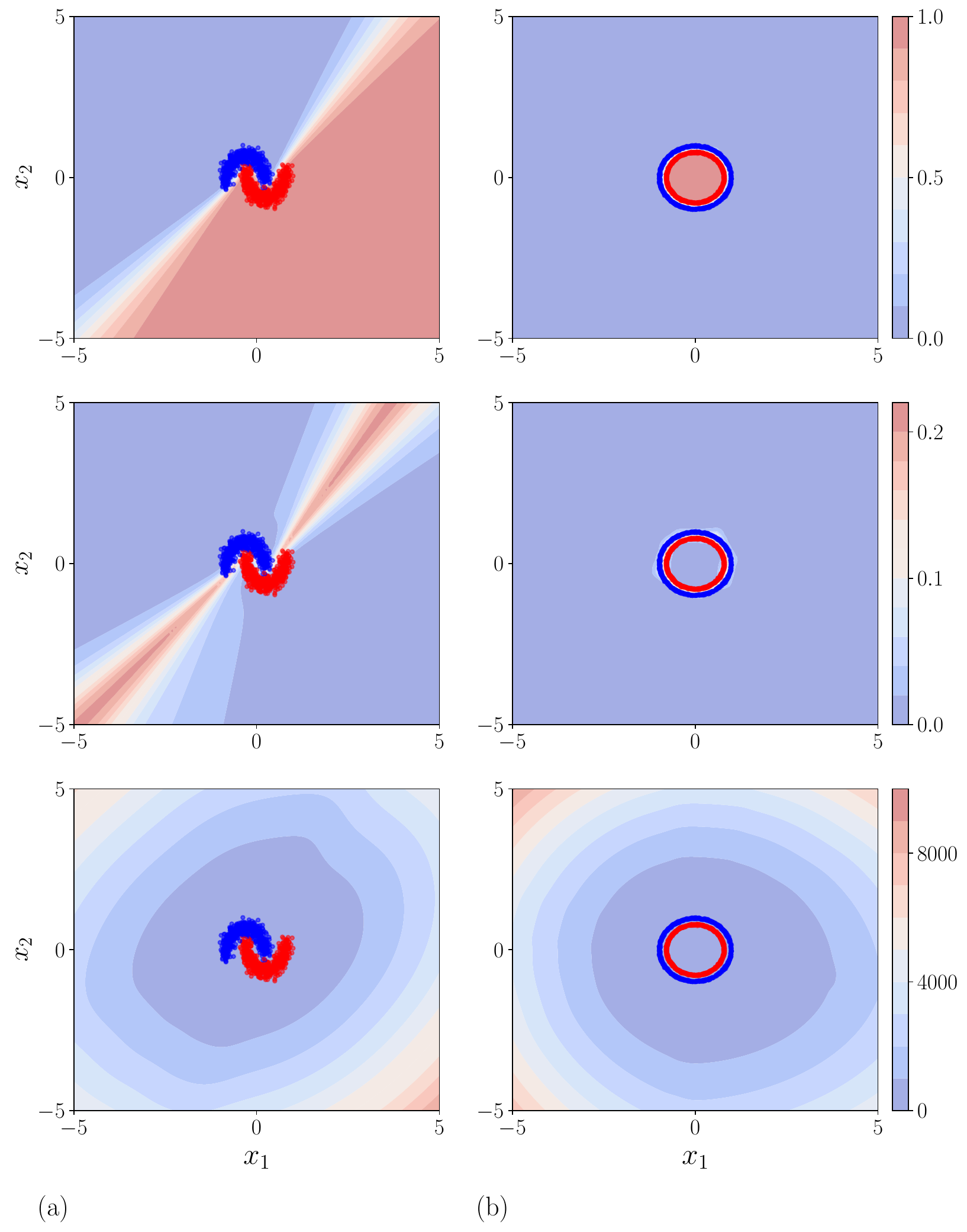}
    \vspace*{-1mm}\caption{Results of the KBNN being trained on data taken from the range $[-1, 1]\times[-1,1]$ for (a) the Moon and (b) Circles datasets. The resulting KBNNs are evaluated on the data range $[-5, 5]\times [-5, 5]$. The first row shows the predictive mean $\mu_y$, the second row the predictive variance $\sigma_{y}^2$, and the third row the variance $(\sigma_{a}^L)^2$ being the variance before the sigmoid activation.}
    \label{fig:unknown}
\end{figure*}

In Figure~\ref{fig:unknown}, we train KBNNs on a small data range $\vx \in [-1, 1] \times [-1, 1]$ for both the Moon and Circles datasets, but evaluate them in much wider data range, to check its behavior on an unknown and never-seen data space. As a comparison, we show the variance of the last layer's output before and after the sigmoid activation, namely $\sigma_{y}^2 = (\sigma_z^{L+1})^2$ and $(\sigma_{a}^L)^2$, respectively. 
As can be seen in Figure~\ref{fig:unknown}, the KBNN returns high uncertainties for $(\sigma_{a}^L)^2$ in areas where it never saw training data, which is as expected according to intuition. However, the uncertainties quantified by $\sigma_{y}^2$ are bounded and tend to go to zero. A possible solution is that we take the variance $(\sigma_{a}^L)^2$ instead of $\sigma_{y}^2$ for quantifying the output uncertainties when using a sigmoid activation for the last layer. An alternative solution could be to resort to a Bernoulli distribution instead of a Gaussian distribution for the output $y$ as proposed in \cite{hennig2020bayesian}.

\begin{figure*}[t]
    \centering
    \includegraphics[width=0.7\textwidth]{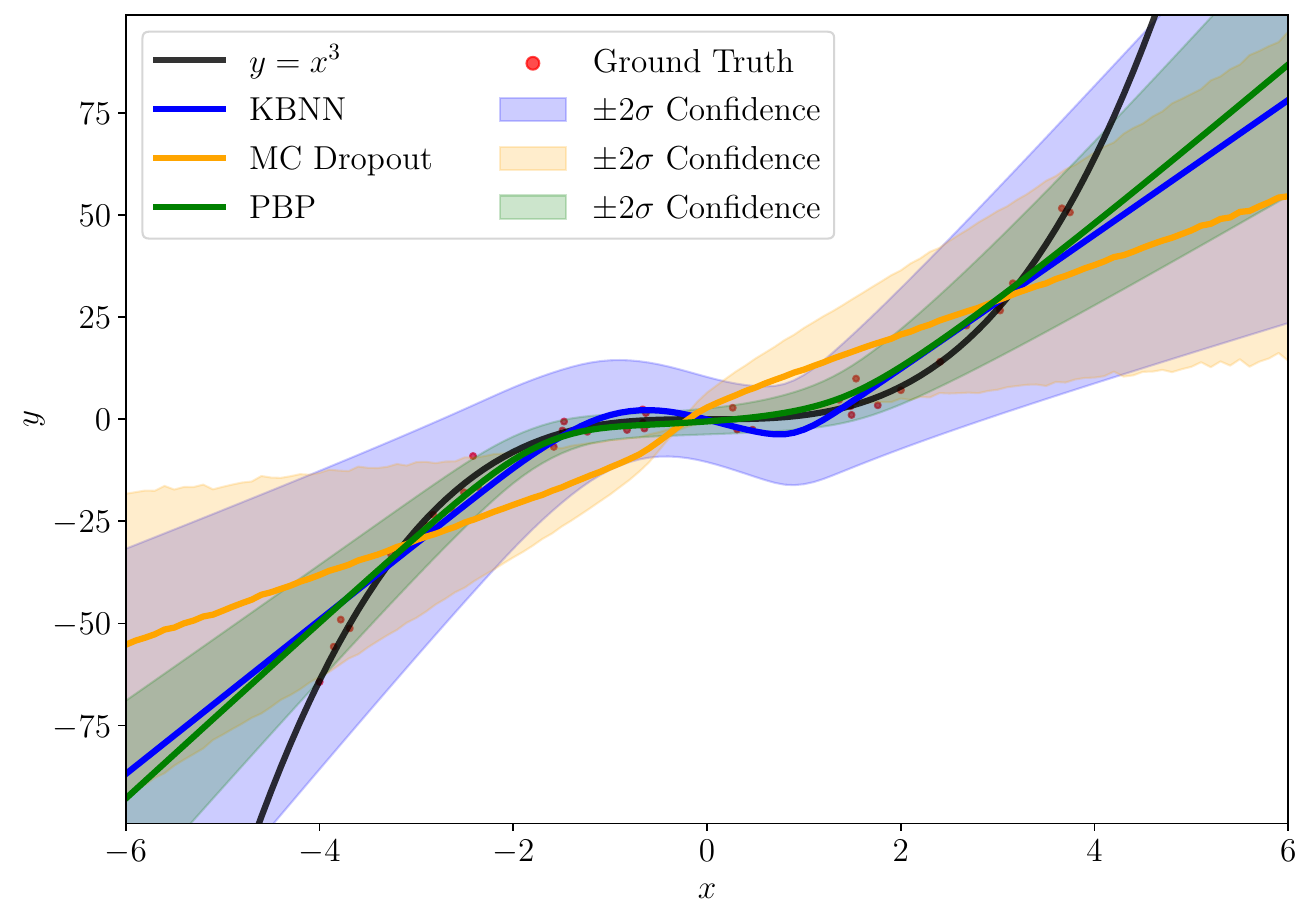}
    \vspace*{-1mm}\caption{Predictions of KBNN, MC Dropout and PBP for the regression tasks $y = x^3 + \epsilon_n$, where $\epsilon_n \sim \SN(0, 9)$, trained on 40 data instances for 20 epochs. }
    \label{fig:toy_regression20}
\end{figure*}

\begin{figure*}[t]
    \centering
    \includegraphics[width=1\textwidth]{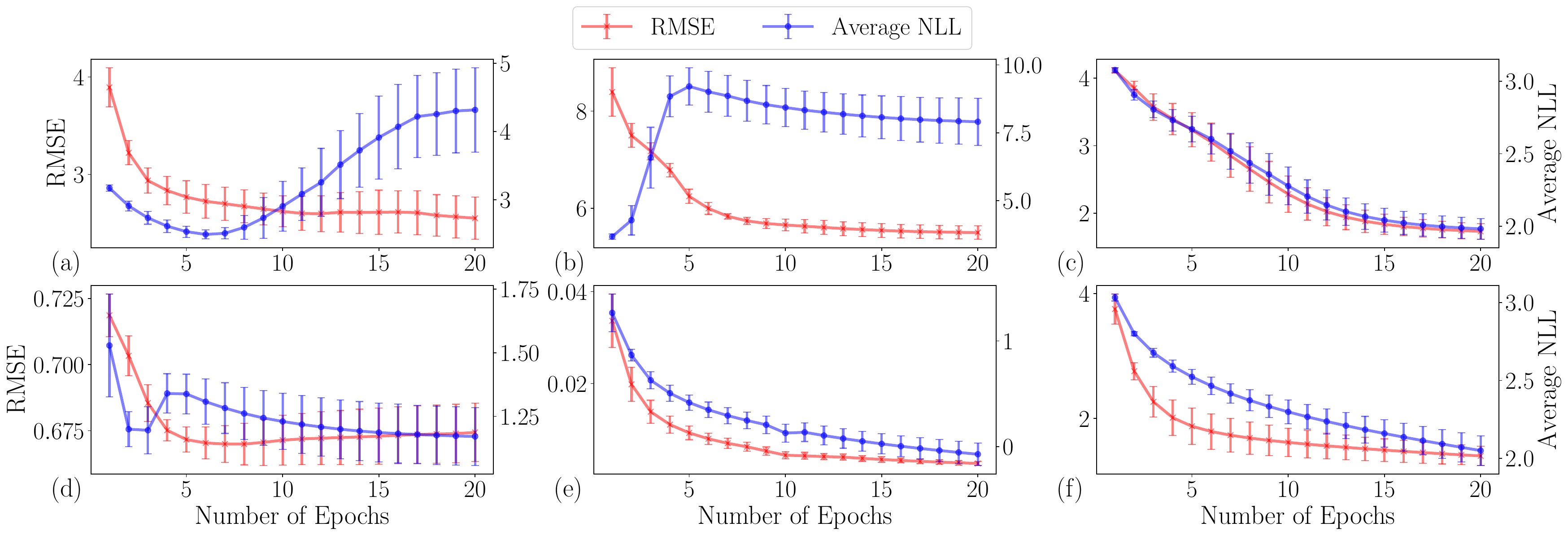}
    \vspace*{-2mm}\caption{The learning progress of KBNN observed for 20 epochs and improvement of RMSE (left axis) and NLL (right axis) on (a) Boston, (b) Concrete, (c) Energy, (d) Wine, (e) Naval, and (f) Yacht datasets. The average RMSE on each dataset is drawn as red line with error bars, while that of average NLL is drawn as blue line.}
    \label{fig:epochs}
\end{figure*}

\section{Synthetic Regression Task on Fewer Data Instances}
\label{sm:synthetic}

In this section we test the learning ability of the KBNN compared to MC Dropout and PBP for a small number of data samples. Figure~\ref{fig:toy_regression20} shows the experiment on the same regression tasks as considered in Sec.~\ref{sec:experiments_regression-synth}. The architecture and initialization of KBNN, MC Dropout, and PBP are as described in Sec.~\ref{sec:experiments_regression-synth}. But instead of 800 training instances, we now trained on 40 instances for 20 epochs. Considering the small data size, we used batch size 2 for MC Dropout.  Compared to MC Dropout, the KBNN provides a significantly better approximation of the ground truth and in addition, provides a reasonable predictive uncertainty. The results of KBNN and PBP are very similar in terms of the predictive mean, while KBNN provides an higher predictive variance.

\section{Learning Progress on the UCI Datasets}
\label{sm:uci}
Fig.~\ref{fig:epochs} shows the learning progress of the KBNN with multiple epochs on the UCI datasets. Most of the time, KBNN is improving with an increasing number of epochs and converges fast. For the Boston, Concrete and Wine datasets, the NLL increases in between. This increase is caused by too low uncertainties in certain areas, where a small deviation between the prediction and ground truth results in larger NLL values. Potential countermeasures to this behavior could be to perform \enquote{noise inflation}, i.e., during each forward pass a Gaussian noise term is added to the weights as suggested in \cite{watanabe1990, puskorius2001parameter}. Also, adding an output noise term as in \cite{hernandez2015probabilistic} would be feasible.

\section{Influence of the Network Architecture}
\label{sm:architecture}

In Table~\ref{table:UCI-performance-layers}--\ref{table:UCI-efficiency-neurons} we show the influence of different KBNN architectures on the regression performance on 
some of 
the considered UCI datasets. For Table~\ref{table:UCI-performance-layers} and Table~\ref{table:UCI-efficiency-layers}, we performed experiments on KBNNs with different numbers of hidden layers, but always with 10 hidden neurons in each layer. For Table~\ref{table:UCI-performance-neurons} and Table~\ref{table:UCI-efficiency-neurons} instead, KBNNs with one hidden layer but a varying number of neurons for this hidden layer are  used. The results indicate that more complicated architectures can lead to a better performance, but after some size of the network, the performance degrades again.

With ten neurons in each hidden layer, KBNNs with two hidden layers obtain best RMSE values in most cases, while KBNNs with three hidden layers achieve best NLL values in the majority of the cases. With one hidden layer but a different number of neurons, KBNNs with 50 neurons achieve the best performance for the most datasets. Considering computational efficiency, the training time increases linearly with an increasing number of hidden layers as Table~\ref{table:UCI-efficiency-layers} indicates. Interestingly, for an increasing number of neurons for a single hidden layer, the training time increases sub-linearly, which is better than expected.

\begin{table*}[t]
\small
\centering
\caption{Performance on UCI datasets for KBNN with different number of hidden layers, with 10 neurons in each hidden layer and trained for 10 epochs.}
\vspace*{-2mm}
\tabcolsep=0.07cm
\scalebox{0.75}{
\begin{tabular}{l cccc cccc}
\hline
&\multicolumn{4}{c}{RMSE} &\multicolumn{4}{c}{NLL} \\\hline
Dataset & 1 layer & 2 layer & 3 layer & 4 layer & 1 layer & 2 layer & 3 layer & 4 layer \\\hline
Boston & $3.085\pm 0.058$ & $2.937\pm 0.267$ & $\mathbf{2.935\pm 0.275}$ & $3.598\pm 0.240$ & 
$28.661\pm 4.692$ & $6.984\pm 1.460$ & $\mathbf{2.643\pm 0.211}$ & $3.788\pm 0.315$ \\
Concrete & $6.656\pm 0.231$ & $\mathbf{5.707\pm 0.233}$ & $5.727\pm 0.363$ & $6.180\pm 0.447$ &
$28.411\pm 4.384$ & $6.663\pm 2.945$ & $4.386\pm 0.603$ & $\mathbf{4.227\pm 0.807}$ \\
Energy & $2.849\pm 0.459$ & $\mathbf{2.151\pm 0.131}$ & $2.742\pm 0.496$ & $3.791\pm 0.422$ &
$17.486\pm 9.349$ & $2.893\pm 0.592$ & $\mathbf{2.648\pm 0.383}$ & $2.984\pm 0.308$ \\
Wine & $0.682\pm 0.006$ & $\mathbf{0.678\pm 0.007}$ & $0.689\pm 0.006$ & $0.733\pm 0.041$ &
$11.677\pm 3.467$ & $2.822\pm 1.022$ & $\mathbf{1.249\pm 0.175}$ & $1.530\pm 0.284$ \\
Naval & $\mathbf{0.003\pm 0.001}$ & $0.004\pm 0.002$ & $0.006\pm 0.004$ & $0.008\pm 0.004$ &
$\mathbf{-1.025\pm 0.356}$ & $-0.623\pm 0.369$ & $-0.213\pm 1.027$ & $-0.479\pm 0.632$\\
Yacht & $1.262\pm 0.180$ & $\mathbf{1.120\pm 0.111}$ & $1.511\pm 0.182$ & $4.115 \pm 0.418$ &
$2.898\pm 0.720$ & $\mathbf{1.200\pm 0.102}$ & $2.285\pm 0.142$ & $4.461\pm 0.038$\\
\hline
\end{tabular}}
\label{table:UCI-performance-layers}
\vspace{3mm}
\centering
\caption{Training time on UCI datasets for KBNN with different number of hidden layers, with 10 neurons in each hidden layer and trained for 10 epochs.}
\vspace*{-2mm}
\small
\tabcolsep=0.1cm
\scalebox{0.9}{
\begin{tabular}{l c c cccc}
\hline
&&&\multicolumn{4}{c}{Training Time / s}\\\hline
Dataset & $\mathit{N}$ & $\mathit{d}$ & 1 layer & 2 layer & 3 layer & 4 layer \\\hline
Boston & $506$ & $13$ & $6.978\pm 0.151$ & $11.665\pm 0.244$ & $16.348\pm 0.512$ & $23.239\pm 3.872$ \\
Concrete & $1,030$ & $8$ & $12.718\pm 0.066$ & $20.695\pm 0.094$ & $28.416\pm 0.088$ & $39.399\pm 1.855$\\
Energy & $768$ & $8$ & $9.515\pm 0.041$ & $15.417\pm 0.044$ & $21.104\pm 0.082$ & $27.068\pm 0.087$\\
Wine & $4,898$ & $11$ & $60.119\pm 0.325$ & $97.037\pm 0.069$  & $133.110\pm 0.335$ & $171.018\pm 0.605$\\
Naval & $11,934$ & $16$ & $179.523\pm 3.820$ & $298.670\pm 3.466$ & $420.540\pm 1.454$ & $484.847\pm 1.845$\\
Yacht & $308$ & $6$ & $4.118\pm 0.069$ & $6.451\pm 0.083$ & $8.817\pm 0.053$ & $11.136\pm 0.054$\\
\hline
\end{tabular}}
\label{table:UCI-efficiency-layers}
\end{table*}
\begin{table*}[t]
\small
\centering
\caption{Performance on UCI datasets for KBNN with different number of hidden neurons, with one hidden layer and trained for 10 epochs.}
\vspace*{-2mm}
\tabcolsep=0.07cm
\scalebox{0.75}{
\begin{tabular}{l cccc cccc}
\hline
&\multicolumn{4}{c}{RMSE} &\multicolumn{4}{c}{NLL} \\\hline
Dataset & 10 neurons & 50 neurons & 100 neurons & 200 neurons & 10 neurons & 50 neurons & 100 neurons & 200 neurons \\\hline
Boston & $3.085\pm 0.058$ & $\mathbf{2.695\pm 0.155}$ & $2.766\pm 0.093$ & $2.840\pm 0.159$ & 
$28.661\pm 4.692$ & $\mathbf{2.767\pm 0.392}$ & $3.036\pm 0.029$ & $3.649\pm 0.025$ \\
Concrete & $6.656\pm 0.231$ & $\mathbf{5.703\pm 0.183}$ & $6.428\pm 0.157$ & $6.441\pm 0.226$ &
$28.411\pm 4.384$ & $8.428\pm 0.946$ & $\mathbf{3.269\pm 0.051}$ & $3.402\pm 0.037$ \\
Energy & $2.849\pm 0.459$ & $\mathbf{2.404\pm 0.259}$ & $3.008\pm 0.255$ & $3.404\pm 0.167$ &
$17.486\pm 9.349$ & $\mathbf{2.394\pm 0.159}$ & $2.899\pm 0.014$ & $3.453\pm 0.019$ \\
Wine & $0.682\pm 0.006$ & $\mathbf{0.666\pm 0.006}$ & $0.673\pm 0.007$ & $0.692\pm 0.009$ &
$11.677\pm 3.467$ & $\mathbf{1.127\pm 0.122}$ & $1.377\pm 0.219$ & $2.500\pm 0.100$ \\
Naval & $\mathbf{0.003\pm 0.001}$ & $0.004\pm 0.002$ & $0.010\pm 0.002$ & $0.030\pm 0.004$ &
$\mathbf{-1.025\pm 0.356}$ & $0.128\pm 0.282$ & $1.002\pm 0.100$ & $2.026\pm 0.198$\\
Yacht & $\mathbf{1.262\pm 0.180}$ & $1.584\pm 0.178$ & $1.662\pm 0.285$ & $2.586 \pm 0.295$ &
$2.898\pm 0.720$ & $\mathbf{2.325\pm 0.055}$ & $2.925\pm 0.026$ & $3.501\pm 0.006$\\
\hline
\end{tabular}}
\label{table:UCI-performance-neurons}
\centering
\vspace{3mm}
\caption{Training time on UCI datasets for KBNN with different number of hidden neurons, with one hidden layer and trained for 10 epochs.}
\vspace*{-2mm}
\small
\tabcolsep=0.1cm
\scalebox{0.9}{
\begin{tabular}{l c c cccc}
\hline
&&&\multicolumn{4}{c}{Training Time / s}\\\hline
Dataset & $\mathit{N}$ & $\mathit{d}$ & 10 neurons & 50 neurons & 100 neurons & 200 neurons \\\hline
Boston & $506$ & $13$ & $6.978\pm 0.151$ & $8.679\pm 0.326$ & $10.208\pm 1.038$ & $10.331\pm 0.412$ \\
Concrete & $1,030$ & $8$ & $12.718\pm 0.066$ & $17.538\pm 0.433$ & $18.459\pm 0.085$ & $20.080\pm 0.252$\\
Energy & $768$ & $8$ & $9.515\pm 0.041$ & $13.224\pm 0.167 $ & $13.569\pm 0.217$ & $15.103\pm 0.308$\\
Wine & $4,898$ & $11$ & $60.119\pm 0.325$ & $86.733\pm 0.886$  & $90.346\pm 1.301$ & $99.994\pm 0.775$\\
Naval & $11,934$ & $16$ & $179.523\pm 3.820$ & $205.297\pm 3.397$ & $221.817\pm 1.377$ & $244.877\pm 0.815$\\
Yacht & $308$ & $6$ & $4.118\pm 0.069$ & $4.955\pm 0.277$ & $5.557\pm 0.139$ & $6.259\pm 0.306$\\
\hline
\end{tabular}}
\label{table:UCI-efficiency-neurons}
\end{table*}

%% file: main.bbl
\begin{thebibliography}{45}
\providecommand{\natexlab}[1]{#1}

\bibitem[{Achille and Soatto(2018)}]{Achille2018}
Achille, A.; and Soatto, S. 2018.
\newblock {Emergence of Invariance and Disentanglement in Deep
  Representations}.
\newblock \emph{Journal of Machine Learning Research}, 19(1): 1947–1980.

\bibitem[{Amisha, Pathania, and Rathaur(2019)}]{amisha2019overview}
Amisha, P.~M.; Pathania, M.; and Rathaur, V.~K. 2019.
\newblock Overview of artificial intelligence in medicine.
\newblock \emph{Journal of family medicine and primary care}, 8(7):
  2328–2331.

\bibitem[{Begoli, Bhattacharya, and Kusnezov(2019)}]{begoli2019need}
Begoli, E.; Bhattacharya, T.; and Kusnezov, D. 2019.
\newblock The need for uncertainty quantification in machine-assisted medical
  decision making.
\newblock \emph{Nature Machine Intelligence}, 1(1): 20--23.

\bibitem[{Bengio(2012)}]{Bengio2012}
Bengio, Y. 2012.
\newblock \emph{{Practical Recommendations for Gradient-Based Training of Deep
  Architectures}}, 437--478.
\newblock Springer Berlin Heidelberg.

\bibitem[{Bingham et~al.(2019)Bingham, Chen, Jankowiak, Obermeyer, Pradhan,
  Karaletsos, Singh, Szerlip, Horsfall, and Goodman}]{bingham2018pyro}
Bingham, E.; Chen, J.~P.; Jankowiak, M.; Obermeyer, F.; Pradhan, N.;
  Karaletsos, T.; Singh, R.; Szerlip, P.; Horsfall, P.; and Goodman, N.~D.
  2019.
\newblock {Pyro: Deep Universal Probabilistic Programming}.
\newblock \emph{Journal of Machine Learning Research}, 20(28): 1--6.

\bibitem[{Dua and Graff(2017)}]{Dua2019uci}
Dua, D.; and Graff, C. 2017.
\newblock {UCI Machine Learning Repository}.

\bibitem[{Duane et~al.(1987)Duane, Kennedy, Pendleton, and
  Roweth}]{DUANE1987hmc}
Duane, S.; Kennedy, A.~D.; Pendleton, B.~J.; and Roweth, D. 1987.
\newblock {Hybrid Monte Carlo}.
\newblock \emph{Physics letters B}, 195(2): 216--222.

\bibitem[{El-Shamouty et~al.(2019)El-Shamouty, Kleeberger, L{\"a}mmle, and
  Huber}]{el2019simulation}
El-Shamouty, M.; Kleeberger, K.; L{\"a}mmle, A.; and Huber, M. 2019.
\newblock Simulation-driven machine learning for robotics and automation.
\newblock \emph{tm-Technisches Messen}, 86(11): 673--684.

\bibitem[{Gal and Ghahramani(2016)}]{gal16dropout}
Gal, Y.; and Ghahramani, Z. 2016.
\newblock {Dropout as a Bayesian Approximation: Representing Model Uncertainty
  in Deep Learning}.
\newblock In \emph{Proceedings of The 33rd International Conference on Machine
  Learning}, Proceedings of Machine Learning Research, 1050--1059.

\bibitem[{Geman and Geman(1984)}]{geman1984gibbssampling}
Geman, S.; and Geman, D. 1984.
\newblock {Stochastic relaxation, Gibbs distributions, and the Bayesian
  restoration of images}.
\newblock \emph{IEEE Transactions on pattern analysis and machine
  intelligence}, (6): 721--741.

\bibitem[{Ghosh, Fave, and Yedidia(2016)}]{Ghosh2016}
Ghosh, S.; Fave, F. M.~D.; and Yedidia, J. 2016.
\newblock {Assumed Density Filtering Methods for Learning Bayesian Neural
  Networks}.
\newblock In \emph{Proceedings of the 30th AAAI Conference on Artificial
  Intelligence}, 1589--1595.

\bibitem[{Graves(2011)}]{graves2011practical}
Graves, A. 2011.
\newblock Practical variational inference for neural networks.
\newblock In \emph{Advances in neural information processing systems},
  2348--2356.

\bibitem[{Hern{\'a}ndez-Lobato and Adams(2015)}]{hernandez2015probabilistic}
Hern{\'a}ndez-Lobato, J.~M.; and Adams, R. 2015.
\newblock Probabilistic backpropagation for scalable learning of bayesian
  neural networks.
\newblock In \emph{International Conference on Machine Learning}, 1861--1869.
  PMLR.

\bibitem[{Hoffman et~al.(2013)Hoffman, Blei, Wang, and
  Paisley}]{JMLR:v14:hoffman13a}
Hoffman, M.~D.; Blei, D.~M.; Wang, C.; and Paisley, J. 2013.
\newblock {Stochastic Variational Inference}.
\newblock \emph{Journal of Machine Learning Research}, 14(4): 1303--1347.

\bibitem[{Hoffman and Gelman(2014)}]{hoffman2014no}
Hoffman, M.~D.; and Gelman, A. 2014.
\newblock {The No-U-Turn sampler: adaptively setting path lengths in
  Hamiltonian Monte Carlo.}
\newblock \emph{Journal of Machine Learning Research.}, 15(1): 1593--1623.

\bibitem[{Huber(2015)}]{huber2015nonlinear}
Huber, M.~F. 2015.
\newblock \emph{{Nonlinear Gaussian Filtering: Theory, Algorithms, and
  Applications}}, volume~19.
\newblock KIT Scientific Publishing.

\bibitem[{Huber(2020)}]{huber2020bayesian}
Huber, M.~F. 2020.
\newblock {Bayesian Perceptron: Towards fully Bayesian Neural Networks}.
\newblock In \emph{59th IEEE Conference on Decision and Control}, 3179--3186.

\bibitem[{Kalman(1960)}]{kalman1960filter}
Kalman, R.~E. 1960.
\newblock {A New Approach to Linear Filtering and Prediction Problems}.
\newblock \emph{Journal of Basic Engineering}, 82(1): 35--45.

\bibitem[{Kingma and Welling(2014)}]{Kingma2014AutoEncodingVB}
Kingma, D.~P.; and Welling, M. 2014.
\newblock {Auto-Encoding Variational Bayes}.
\newblock \emph{CoRR}, abs/1312.6114.

\bibitem[{Kirkpatrick et~al.(2017)Kirkpatrick, Pascanu, Rabinowitz, Veness,
  Desjardins, Rusu, Milan, Quan, Ramalho, Grabska-Barwinska, Hassabis, Clopath,
  Kumaran, and Hadsell}]{kirkpatrick2017forgetting}
Kirkpatrick, J.; Pascanu, R.; Rabinowitz, N.; Veness, J.; Desjardins, G.; Rusu,
  A.~A.; Milan, K.; Quan, J.; Ramalho, T.; Grabska-Barwinska, A.; Hassabis, D.;
  Clopath, C.; Kumaran, D.; and Hadsell, R. 2017.
\newblock Overcoming catastrophic forgetting in neural networks.
\newblock \emph{Proceedings of the National Academy of Sciences}, 114(13):
  3521--3526.

\bibitem[{Kristiadi, Hein, and Hennig(2020)}]{hennig2020bayesian}
Kristiadi, A.; Hein, M.; and Hennig, P. 2020.
\newblock {Being Bayesian, Even Just a Bit, Fixes Overconfidence in ReLU
  Networks}.
\newblock In \emph{Proceedings of the 37th International Conference on Machine
  Learning}, volume 119, 5436--5446.

\bibitem[{Kurle et~al.(2019)Kurle, Cseke, Klushyn, van~der Smagt, and
  G{\"u}nnemann}]{kurle2019continual}
Kurle, R.; Cseke, B.; Klushyn, A.; van~der Smagt, P.; and G{\"u}nnemann, S.
  2019.
\newblock Continual learning with bayesian neural networks for non-stationary
  data.
\newblock In \emph{International Conference on Learning Representations}.

\bibitem[{Lu, Ie, and Sha(2021)}]{Lu2021}
Lu, Z.; Ie, E.; and Sha, F. 2021.
\newblock {Mean-Field Approximation to Gaussian-Softmax Integralwith
  Application to Uncertainty Estimation}.
\newblock \emph{arXiv preprint arXiv:2006.07584v2}.

\bibitem[{MacKay(1992)}]{mackay1992practical}
MacKay, D.~J. 1992.
\newblock A practical Bayesian framework for backpropagation networks.
\newblock \emph{Neural computation}, 4(3): 448--472.

\bibitem[{Maybeck(1979)}]{Maybeck1979}
Maybeck, P.~S. 1979.
\newblock \emph{Stochastic {Models, Estimation, and Control}}, volume 141 of
  \emph{Mathematics in Science and Engineering}.
\newblock Academic Press, Inc.

\bibitem[{Metropolis et~al.(1953)Metropolis, Rosenbluth, Rosenbluth, Teller,
  and Teller}]{Metropolis1953equationOS}
Metropolis, N.; Rosenbluth, A.~W.; Rosenbluth, M.; Teller, A.~H.; and Teller,
  E. 1953.
\newblock Equation of state calculations by fast computing machines.
\newblock \emph{Journal of Chemical Physics}, 21: 1087--1092.

\bibitem[{Minka(2001)}]{minka2001}
Minka, T. 2001.
\newblock \emph{{A family of algorithms for approximate Bayesian inference}}.
\newblock Ph.D. thesis, USA.

\bibitem[{Minka, Xiang, and Qi(2009)}]{minka2009vvm}
Minka, T.~P.; Xiang, R.; and Qi, Y.~A. 2009.
\newblock Virtual Vector Machine for Bayesian Online Classification.
\newblock In \emph{Proceedings of the Twenty-Fifth Conference on Uncertainty in
  Artificial Intelligence}, UAI '09, 411–418. Arlington, Virginia, USA: AUAI
  Press.
\newblock ISBN 9780974903958.

\bibitem[{Murphy(2012)}]{murphy2012machine}
Murphy, K.~P. 2012.
\newblock \emph{Machine learning: a probabilistic perspective}.
\newblock MIT press.

\bibitem[{Neal(1995)}]{neal1995bnn}
Neal, R.~M. 1995.
\newblock \emph{{Bayesian Learning for Neural Networks}}.
\newblock Ph.D. thesis, CAN.
\newblock AAINN02676.

\bibitem[{Nguyen et~al.(2018)Nguyen, Li, Bui, and Turner}]{Nguyen2018}
Nguyen, C.~V.; Li, Y.; Bui, T.~D.; and Turner, R.~E. 2018.
\newblock {Variational Continual Learning}.
\newblock In \emph{International Conference on Learning Representations}.

\bibitem[{Opper(1998)}]{Opper1998}
Opper, M. 1998.
\newblock A {Bayesian Approach to Online Learning}.
\newblock In \emph{Online Learning in Neural Networks}, 363--378.

\bibitem[{Parisi et~al.(2019)Parisi, Kemker, Part, Kanan, and
  Wermter}]{parisi2019}
Parisi, G.~I.; Kemker, R.; Part, J.~L.; Kanan, C.; and Wermter, S. 2019.
\newblock {Continual lifelong learning with neural networks: A review}.
\newblock \emph{Neural Networks}, 113: 54 -- 71.

\bibitem[{Pedregosa et~al.(2011)Pedregosa, Varoquaux, Gramfort, Michel,
  Thirion, Grisel, Blondel, Prettenhofer, Weiss, Dubourg
  et~al.}]{pedregosa2011scikit}
Pedregosa, F.; Varoquaux, G.; Gramfort, A.; Michel, V.; Thirion, B.; Grisel,
  O.; Blondel, M.; Prettenhofer, P.; Weiss, R.; Dubourg, V.; et~al. 2011.
\newblock {Scikit-learn: Machine learning in Python}.
\newblock \emph{Journal of Machine Learning Research}, 12: 2825--2830.

\bibitem[{Puskorius and Feldkamp(2001)}]{puskorius2001parameter}
Puskorius, G.~V.; and Feldkamp, L.~A. 2001.
\newblock {Parameter-based Kalman Filter Training: Theory and Implementation}.
\newblock \emph{Kalman filtering and neural networks}, 23.

\bibitem[{Rauch, Tung, and Striebel(1965)}]{rauch1965smoother}
Rauch, H.~E.; Tung, F.; and Striebel, C.~T. 1965.
\newblock Maximum likelihood estimates of linear dynamic systems.
\newblock \emph{AIAA Journal}, 3(8): 1445--1450.

\bibitem[{Ritter, Botev, and Barber(2018{\natexlab{a}})}]{ritter2018laplace}
Ritter, H.; Botev, A.; and Barber, D. 2018{\natexlab{a}}.
\newblock {A Scalable Laplace Approximation for Neural Networks}.
\newblock In \emph{Proceedings of the 6th International Conference on Learning
  Representations}.

\bibitem[{Ritter, Botev, and Barber(2018{\natexlab{b}})}]{ritter2018online}
Ritter, H.; Botev, A.; and Barber, D. 2018{\natexlab{b}}.
\newblock Online Structured Laplace Approximations for Overcoming Catastrophic
  Forgetting.
\newblock In \emph{Proceedings of the 32nd International Conference on Neural
  Information Processing Systems}, NIPS'18, 3742–3752. Red Hook, NY, USA:
  Curran Associates Inc.

\bibitem[{Snoek et~al.(2015)Snoek, Rippel, Swersky, Kiros, Satish, Sundaram,
  Patwary, Prabhat, and Adams}]{Snoek2015}
Snoek, J.; Rippel, O.; Swersky, K.; Kiros, R.; Satish, N.; Sundaram, N.;
  Patwary, M. M.~A.; Prabhat, P.; and Adams, R.~P. 2015.
\newblock {Scalable Bayesian Optimization Using Deep Neural Networks}.
\newblock In \emph{Proceedings of the 32nd International Conference on
  International Conference on Machine Learning}, 2171--2180.

\bibitem[{Srivastava et~al.(2014)Srivastava, Hinton, Krizhevsky, Sutskever, and
  Salakhutdinov}]{srivastava2014dropout}
Srivastava, N.; Hinton, G.; Krizhevsky, A.; Sutskever, I.; and Salakhutdinov,
  R. 2014.
\newblock {Dropout: A Simple Way to Prevent Neural Networks from Overfitting}.
\newblock \emph{Journal of Machine Learning Research}, 15(56): 1929--1958.

\bibitem[{Särkkä(2013)}]{saerkkae_2013}
Särkkä, S. 2013.
\newblock \emph{{Bayesian Filtering and Smoothing}}.
\newblock Institute of Mathematical Statistics Textbooks. Cambridge University
  Press.

\bibitem[{Watanabe and Tzafesta(1990)}]{watanabe1990}
Watanabe, K.; and Tzafesta, S.~G. 1990.
\newblock {Learning Algorithms for Neural Networks with the Kalman Filter}.
\newblock \emph{Journal of Intelligent and Robotic Systems}, 3: 305--319.

\bibitem[{Wu et~al.(2019)Wu, Nowozin, Meeds, Turner, Hern\'andez-Lobato, and
  Gaunt}]{wu2019}
Wu, A.; Nowozin, S.; Meeds, E.; Turner, R.~E.; Hern\'andez-Lobato, J.~M.; and
  Gaunt, A.~L. 2019.
\newblock {Deterministic Variational Inference for Robust Bayesian Neural
  Networks}.
\newblock In \emph{Proceedings of the 7th International Conference on Learning
  Representations}.

\bibitem[{Zeng et~al.(2020)Zeng, Song, Lee, Rodriguez, and
  Funkhouser}]{zeng2020tossingbot}
Zeng, A.; Song, S.; Lee, J.; Rodriguez, A.; and Funkhouser, T. 2020.
\newblock {Tossingbot: Learning to Throw Arbitrary Objects with Residual
  Physics}.
\newblock \emph{IEEE Transactions on Robotics}, 36(4): 1307--1319.

\bibitem[{Zhang et~al.(2019)Zhang, Bütepage, Kjellström, and
  Mandt}]{Zhang2019}
Zhang, C.; Bütepage, J.; Kjellström, H.; and Mandt, S. 2019.
\newblock {Advances in Variational Inference}.
\newblock \emph{IEEE Transactions on Pattern Analysis and Machine
  Intelligence}, 41(8): 2008--2026.

\end{thebibliography}
